\documentclass[runningheads]{llncs}
\usepackage{graphicx}
\usepackage{tikz}
\usepackage{comment}
\usepackage{amsmath,amssymb} 
\usepackage{color}
\usepackage[accsupp]{axessibility}  

\usepackage{booktabs}
\usepackage{epsfig}
\usepackage{amsmath}
\usepackage{amssymb}
\usepackage[utf8]{inputenc} 
\usepackage[T1]{fontenc}    
\usepackage{url}            
\usepackage{booktabs}       
\usepackage{amsfonts}       
\usepackage{nicefrac}       
\usepackage{microtype}      
\usepackage{graphicx}
\usepackage{comment}
\usepackage{amsmath,amssymb} 
\usepackage{color}
\usepackage{microtype}
\usepackage{appendix}
\usepackage{xcolor}
\usepackage{mathtools}
\usepackage{multirow}
\usepackage{fixltx2e}
\usepackage{pifont}
\usepackage[procnames]{listings}
\usepackage{wrapfig}
\usepackage{dirtytalk}
\usepackage[ruled,vlined]{algorithm2e}
\usepackage{algorithmic}
\usepackage{subfigure}
\usepackage{appendix}
\newcommand{\methodname}{SITTA}

\newcommand{\textureencodername}{\text{$En_T$}}
\newcommand{\contentencodername}{\text{$En_C$}}
\newcommand{\decoderA}{\text{$De_A$}}
\newcommand{\decoderB}{\text{$De_B$}}
\newcommand{\disA}{\text{$D_A$}}
\newcommand{\disB}{\text{$D_B$}}

\definecolor{bleudefrance}{rgb}{0.0, 0.36, 1.0}

\usepackage[pagebackref,breaklinks,colorlinks]{hyperref}
\usepackage[capitalize]{cleveref}
\crefname{section}{Sec.}{Secs.}
\Crefname{section}{Section}{Sections}
\Crefname{table}{Table}{Tables}
\crefname{table}{Tab.}{Tabs.}

\begin{document}
\pagestyle{headings}
\mainmatter
\def\ECCVSubNumber{11}  

\title{SITTA: Single Image Texture Translation\\ 
for Data Augmentation}

\titlerunning{SITTA: Single Image Texture Translation\\ 
for Data Augmentation}
%
\author{
Boyi Li\inst{1}
\and
Yin Cui\inst{2}
\and
Tsung-Yi Lin\inst{3}
\and
Serge Belongie\inst{4}
}
\authorrunning{B Li et al.}
%
\institute{
Cornell University, Cornell Tech 
\and
Google Research
\and
NVIDIA
\and
University of Copenhagen
}
\maketitle

\begin{abstract}
Recent advances in data augmentation enable one to translate images by learning the mapping between a source domain and a target domain. Existing methods tend to learn the distributions by training a model on a variety of datasets, with results evaluated largely in a subjective manner. Relatively few works in this area, however, study the potential use of image synthesis methods for recognition tasks. In this paper, we propose and explore the problem of image translation for data augmentation. We first propose a lightweight yet efficient model for translating texture to augment images based on a single input of source texture, allowing for fast training and testing, referred to as Single Image Texture Translation for data Augmentation (\methodname{}). Then we explore the use of augmented data in long-tailed and few-shot image classification tasks. We find the proposed augmentation method and workflow is capable of translating the texture of input data into a target domain, leading to consistently improved image recognition performance. Finally, we examine how \methodname{} and related image translation methods can provide a basis for a data-efficient, ``augmentation engineering'' approach to model training. Codes are available at \href{https://github.com/Boyiliee/SITTA}{https://github.com/Boyiliee/SITTA}.

\end{abstract}

\vspace{-0.1in}
\section{Introduction}
\label{sec:intro}
{\small{\say{The Forms are not limited to geometry. For any conceivable thing or property there is a corresponding Form, a perfect example of that thing or property. The list is almost inexhaustible.} 

\quad\quad\quad\quad\quad\quad\quad\quad\quad\quad\quad\quad\quad\quad\quad\quad\quad\quad\quad\quad\quad\quad     ---Plato “Theory of Forms”}}

\begin{wrapfigure}{R}{0.5\linewidth}
    \centering
    \includegraphics[width=0.8\linewidth]{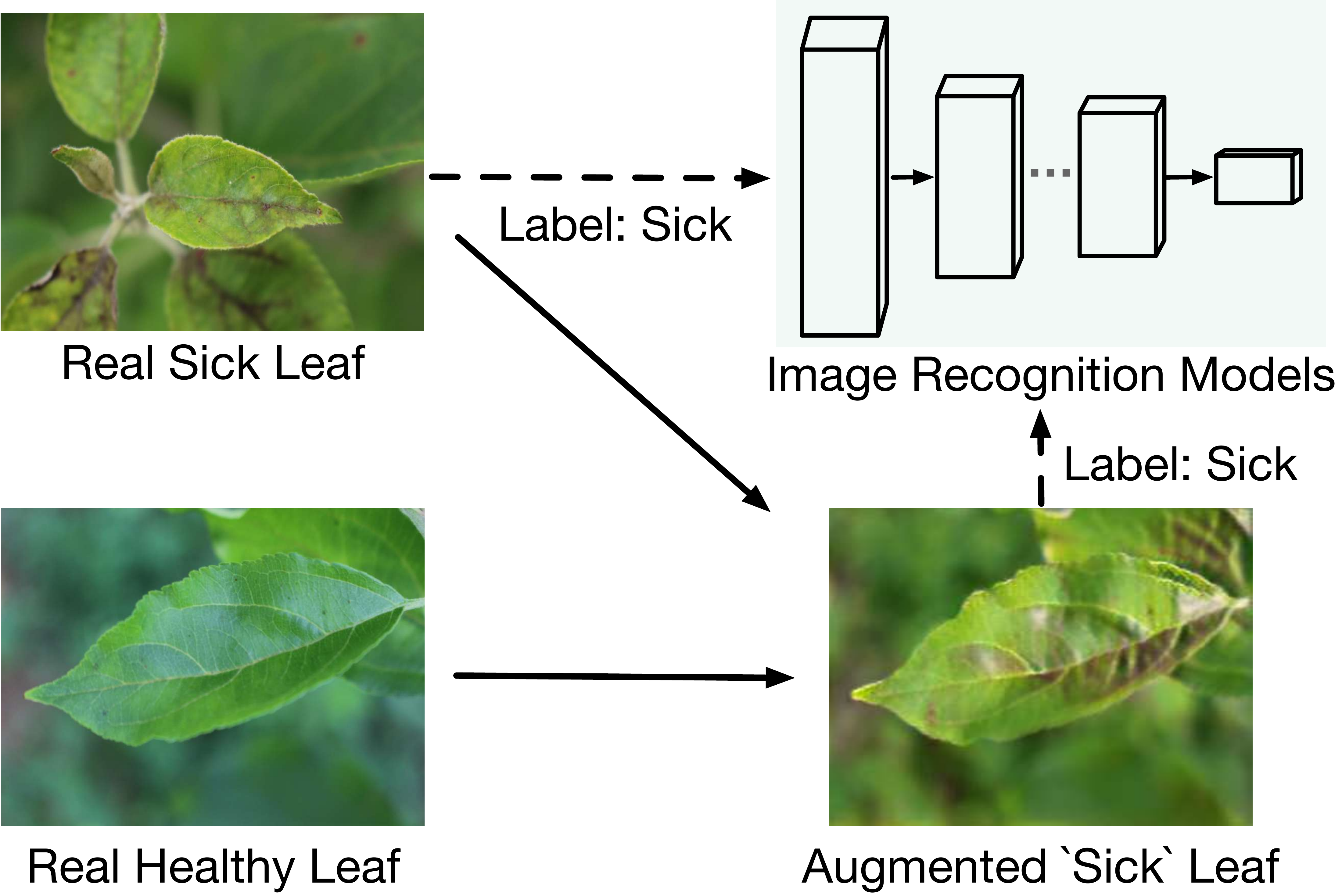}
    \caption{Single image texture translation for data augmentation (\methodname{}) workflow. We first obtain textures from sick leaf and content from healthy leaf, then we feed them into \methodname{} model to generate augmented data. We set the corresponding label as `sick.'}
    \label{fig:sitta_fig1}
    \vspace{-0.2in}
\end{wrapfigure}

Recent years have witnessed a breakthrough in deep learning based image synthesis such as image translation~\cite{gatys2016image,legendre2019deeplight,mildenhall2020nerf} that manipulates or synthesizes images using neural networks rather than hand-crafted techniques, such as guided filtering~\cite{he2012guided} or image quilting~\cite{efros2001image}. However, few of them study the potential use of semantic image synthesis methods as an effective data augmentation tool for recognition tasks. The progress is primarily limited by two bottlenecks: the validity of synthetic data for target labels and the running time. With regard to data validity, many works such as Stylized-ImageNet~\cite{geirhos2018imagenet} propose to apply style transfer to the original dataset for pre-training to improve the model's robustness, but the synthetic images lack the natural appearance of the original images. Also, since it deconstructs texture and content, it will hurt the recognition performance if trained with the augmented data, and needs to be fine-tuned using only the original dataset to obtain a benefit. Though many approaches~\cite{park2020swapping,CycleGAN2017,isola2017image} have proposed advanced algorithms to translate style and texture, most of them still focus on a subjective evaluation. With regard to running time, current techniques~\cite{shaham2019singan,lin2020tuigan} usually need to train for at least several hours. This seriously hinders its use for data augmentation in real applications. While in the domain of data augmentation, current methods mainly focus on pixel-level or geometric operations such as blur or crop~\cite{shorten2019survey}. The potential of augmenting data into different domains, however, is relatively unexplored.

\begin{wrapfigure}{R}{0.99\linewidth}
    \centering
    \includegraphics[width=\linewidth]{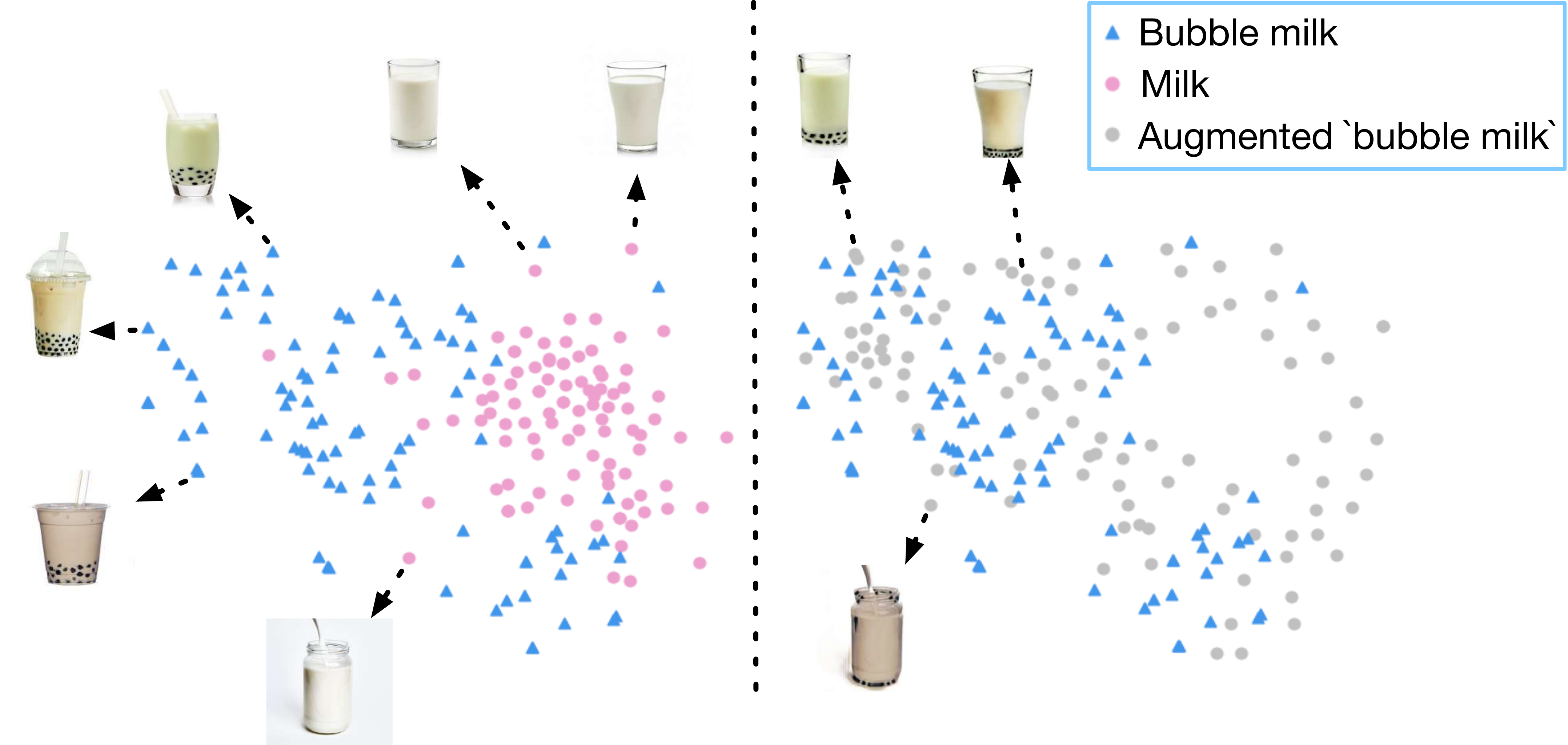}
    \vspace{-0.1in}
    \caption{t-SNE embedding before and after using \methodname{}. Left: milk (red) and bubble milk (blue) images. Right: corresponding augmented `bubble milk' (gray) and original bubble milk images.}
    \label{fig:augE_sitt_tsne}
    \vspace{-0.3in}
\end{wrapfigure}

In light of this, we propose to explore, design and study an efficient Single Image Texture Translation for data Augmentation (\methodname{}) in image recognition tasks. It enables texture translation or swapping trained with one-shot texture source input. We believe an ideal image synthesis method for data augmentation should yield visually appealing results, improved recognition performance as well as time efficiency. \methodname{} is the first of few methods that explore this problem and try to balance all these factors. Our model is lightweight and permits fast training ($<$ 5 min) and testing (9 ms) on a $288 \times 288$ image using a single Geforce GTX 1080 Ti GPU.
In Figure~\ref{fig:augE_sitt_tsne}, we illustrate an example of generated `bubble milk' images by translating the texture of bubble milk to milk images. 
We visualize the image distribution using t-SNE~\cite{maaten2008visualizing}. We collect 100 natural milk images and 100 natural bubble milk images from the Web, shown on the left. 
On the right, we can see that the \methodname{} `bubble milk' images align well with the original bubble milk images, which hints the efficacy of \methodname{} for semantic data augmentation. 
The intuition of replacing texture is conceptually similar to the arithmetic properties of word embeddings~\cite{bolukbasi2016man}, e.g., $\overrightarrow{Milk} - \overrightarrow{\text{texture}(Milk)} + \overrightarrow{\text{texture}(Bubble Milk)} = \overrightarrow{Bubble Milk}$. 

Our key contributions in this work are as follows.
\begin{itemize}
  \item[1.]  We introduce a new lightweight single image texture translation method for data augmentation (\methodname{}) that translates textures from a single image to another.
  \item[2.] 
  We explore the use of \methodname{} for augmenting data towards the target texture domain. As an example, we translate texture from rare diseased leaves to abundant healthy leaves to augment training data and improve image classification results in the plant pathology challenge dataset~\cite{thapa2020plant}.
  \item[3.] We dig deeper into the use of \methodname{} for augmenting data and demonstrate its effectiveness in several image recognition tasks (e.g, improving the performance of ResNet-18 by 1.4\% for 5-shot classification on CUB-200-2011), which sheds light on the potential direction of image synthesis for data augmentation in the wild.
\end{itemize}

\section{Related Work}
\label{sec:related_work}

\textbf{Texture Translation}. Deep learning based image synthesis manipulates images through the design of various generative neural networks. Fueled by the explorations in generative models such as GAN~\cite{goodfellow2014generative} and VAE~\cite{kingma2013auto}, researchers have explored ideas such as neural style transfer and image translation. A neural algorithm of artistic style (ArtStyle)~\cite{gatys2016image} can separate and recombine the image content and style of natural images. CycleGAN~\cite{CycleGAN2017} investigates the use of conditional adversarial networks as a general-purpose solution to image-to-image translation problems. SinGAN~\cite{shaham2019singan} demonstrates success in single image retargeting. Few-Shot Unsupervised Image-to-Image Translation (FUNIT)~\cite{liu2019few} focuses on previously unseen target classes that are specified at test time only by a few example images. TuiGAN~\cite{lin2020tuigan} aims to learn versatile image-to-image translation with two unpaired images designed in a coarse-to-fine manner. Furthermore,~\cite{yoo2019photorealistic,park2020swapping,park2020cut} propose elegant methods for better preserving their structural information and statistical properties. However, these methods focus more on a subjective manner instead of considering all factors including decent outputs, time efficiency, improved recognition results, as well as flexibility with various input resolutions for data augmentation. 

\begin{figure*}
    \centering
    \includegraphics[width=\linewidth]{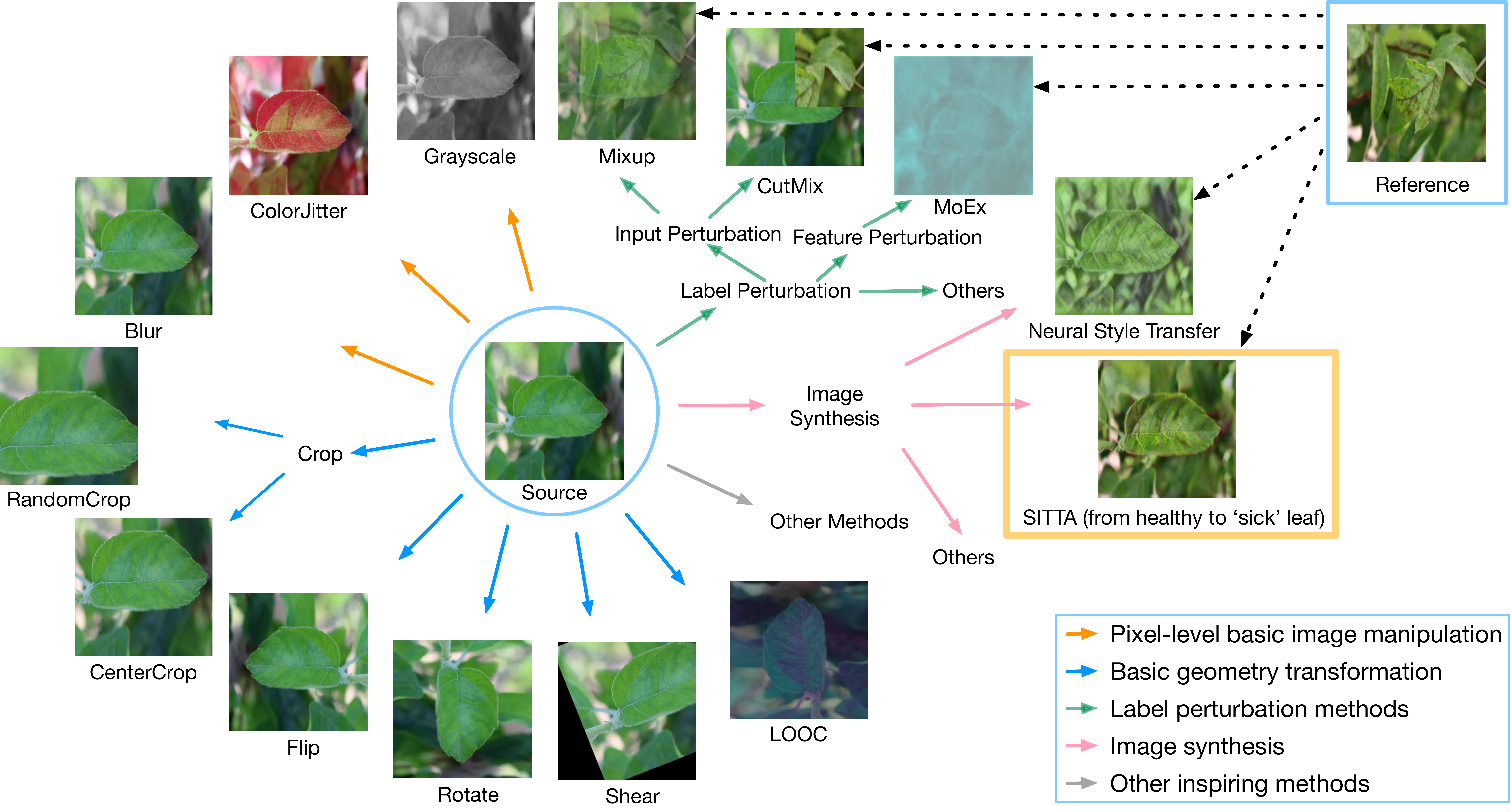}
    \vspace{-0.15in}
    \caption{Overview of the data augmentation landscape.}
    \label{fig:augE}
    \vspace{-0.4in}
\end{figure*}

\textbf{Data Augmentation.}
Data augmentation plays a critical role in various image recognition tasks~\cite{chen2020improved}. On the one hand, basic image manipulation methods have been widely used as effective pre-processing tools~\cite{shorten2019survey,shrivastava2017learning,geirhos2018imagenet,wang2019implicit,li2021feature,he2020momentum,chen2020simple}, with examples including flipping, rotation, jittering, grayscale, and Gaussian blur. Several works~\cite{touvron2019fixing,cubuk2019autoaugment} explore effective combined choices of basic image manipulations. LOOC~\cite{xiao2021what} searches for an optimal augmentation strategy among a large pool of candidates based on specific datasets or tasks. Mixup~\cite{zhang2018mixup} interpolates two training inputs in feature and label space simultaneously. CutMix~\cite{yun2019cutmix} uses a copy-paste strategy and mixes the labels in proportion to the number of pixels contributed by each input image to the final composition. MoEx~\cite{li2021feature} exchange the moments of the learned features of one training image by those of another, and also interpolate the target labels. On the other hand, many researchers have begun to explore the impact of semantic data augmentation for real-world image recognition tasks~\cite{zhang2016colorful,shrivastava2017learning,wang2019implicit,antoniou2017data}. Geirhos, et al~\cite{geirhos2018imagenet} find that CNNs like ResNet-50 favor texture rather than shape, and appropriate use of augmented ImageNet via style transfer could alleviate this bias and improve the model performance. 
In addition,~\cite{zheng2019stada,somavarapu2020frustratingly,zhang2020learning,qin2020gan} aim to learn robust shape-based features for domain generalization. RL-CycleGAN~\cite{rao2020rl} introduces a consistency loss for simulation-to-real-world transfer for reinforcement learning. However, few of them explicitly augment data with natural outputs that could be used to solve real problems. In this paper, we propose \methodname{} that generates new data semantically via domain-specific texture translation. We provide a simplified illustration of the data augmentation landscape in Figure~\ref{fig:augE}.

\vspace{-0.1in}
\section{Method}
\label{sec:method}
\subsection{Workflow}
\methodname{} aims to augment images that align with the target texture domain and could be used as extra training data for various recognition tasks. Assume we have plenty of source domain images SetA but a few target domain images SetB, the only difference between SetA and SetB is that they contain different textures. Our goal is to synthesize extra `B` data by replacing textures of SetA with the new textures from SetB, we call the augmented dataset as AugSetB. Based on \methodname{}, we have two ways to achieve this: i) \textit{Single to Single}. Since \methodname{} enables translating textures between two single images, we train and generate new image $I'_B$ based on every input $I_A$ and $I_B$ from SetA and SetB, respectively.  ii) \textit{Single to Multi}. We train and generate new image $I'_B$ with a single source texture image $I_B$ from SetB and all content images from SetA. Please see Section \textcolor{red}{A} in the Appendix for workflow details.

\subsection{Model Design} 
\begin{figure*}
    \centering
    \includegraphics[width=\linewidth]{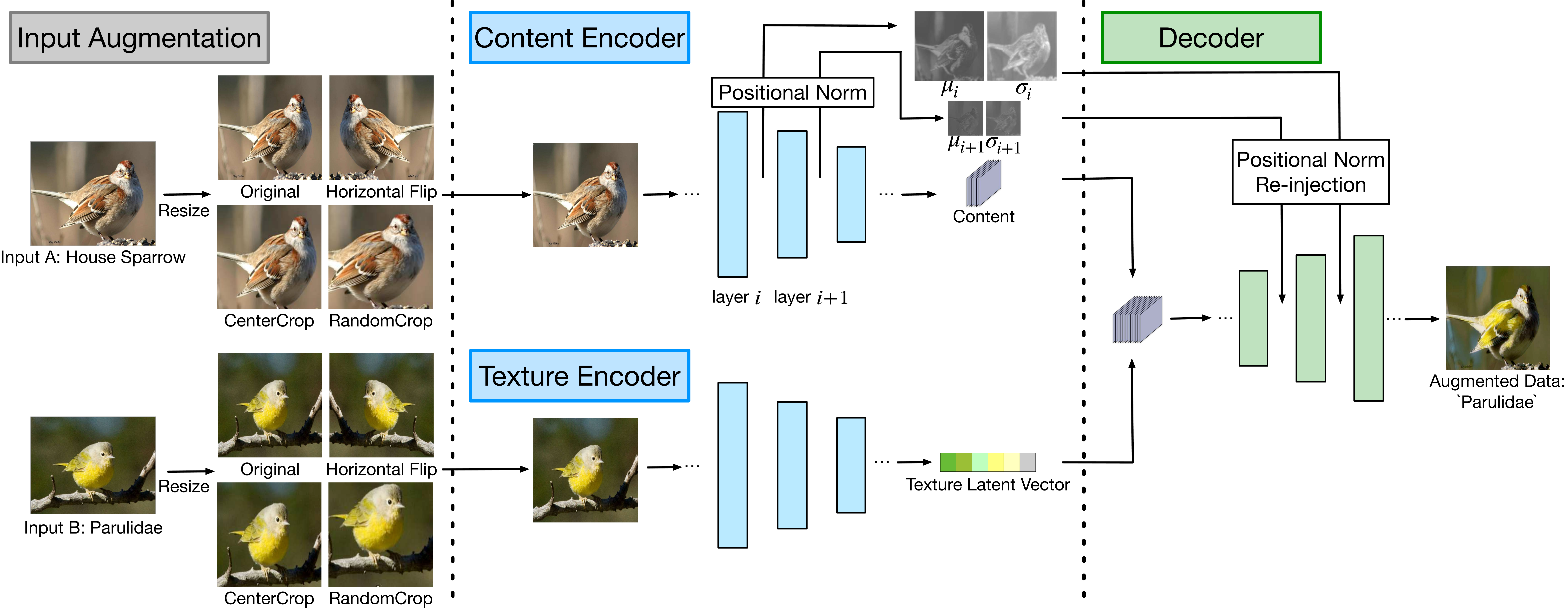}
    \caption{\methodname{} Framework. \methodname{} learns to encode the content and texture from a single content and texture image respectively, then decodes an output image with the texture translated onto the content.}
    \label{fig:framework}
    \vspace{-0.2in}
\end{figure*}

Figure~\ref{fig:framework} shows a brief sketch of our framework. In the illustrated example, \methodname{} aims to translate texture from a single source texture image (Input B: Parulidae) to a content image (Input A: House Sparrow) and get a synthetic `Parulidae' image. 
\methodname{} consists of three parts: shared Texture Encoder (\textureencodername{}), shared Content Encoder (\contentencodername{}) and corresponding Decoder \decoderA{} and \decoderB{} for input A and B. We first feed input B into \textureencodername{} to obtain the texture latent vector and feed input A into \contentencodername{} to get the content (structure) matrix. Then we concatenate the texture and content features and feed them into Decoder to generate the output image.

To efficiently preserve structure and extract textures with an optimal number of layers, we apply two downsampling operations during training. To achieve efficient model design, we emphasize three critical components: input augmentation, structural information re-injection, texture latent regression. 

(1) Input augmentation. We propose to directly augment input by Horizontal Flip, CenterCrop, and RandomCrop.
We find such input augmentation could effectively guarantee the diversity of image scales for fast and efficient training.

(2) Structural information re-injection. To better preserve the structural information and optimize training, we apply Positional Norm~\cite{li2019positional} in \contentencodername{} to extract intermediate normalization constants mean $\mu$ and standard deviation $\sigma$ as structural features $\beta$ and $\gamma$ and re-inject them into the later layers of Decoder to transfer structural information. Given the activations $X \in \mathbb{R}^{B \times C \times H \times W}$ (where $B$ denotes the batch size, $C$ the number of channels, $H$ the height, and $W$ the width) in a given layer of a neural net, the extracting operations\footnote{The $\epsilon$ is a small stability constant (\emph{e.g.}, $\epsilon=10^{-5}$) to avoid divisions by zero and imaginary values due to numerical inaccuracies.} are listed as:
\begin{equation}
\beta = \mu_{b, h, w} = \frac{1}{C}\sum_{c=1}^C X_{b, c, h, w},
\end{equation}
\begin{equation}
\gamma = \sigma_{b, h, w} = \sqrt{ \frac{1}{C}\sum_{c=1}^C \left(X_{b, c, h, w} - \mu_{b, h, w}\right)^2 + \epsilon}.
\end{equation}
The re-injecting operation after $i$th intermediate layer is listed as: 
\begin{equation}
\mathrm{Out_i}(\mathbf{x}) = \gamma_i F_i(\mathbf{x}) + \beta_i,
\end{equation}
where the function $F$ is modeled by the intermediate layers. We summarize the \methodname{} model design in Appendix Section \textcolor{red}{B}.

(3) Texture latent regression. To achieve better representation disentanglement~\cite{DRIT} for texture and content, we use a latent regression loss $L_{idt}$ to encourage the invertible mapping between the latent texture vectors and the corresponding outputs and enforce the reconstruction based on the latent texture vectors.

\subsection{Loss Function}
\methodname{} is based on GAN framework~\cite{goodfellow2014generative}, the loss function is composed of 5 parts: adversarial loss $L_{adv}$, latent regression loss $L_{idt}$, content matrix reconstruction loss $L_{rec}$, texture vector KL divergence loss $L_{kl}$ and perceptual loss $L_{f}$. During training, we have inputs $I_A$ and $I_B$, extracted texture vectors $T_A$ and $T_B$, content matrix $C_A$ and $C_B$, corresponding normalization constants $\beta_A$, $\gamma_A$ and $\beta_B$, $\gamma_B$, as well as outputs $I'_A$ and $I'_B$. 

Adversarial loss is GAN standard loss function for matching the distribution of translated image to the target domain:
\begin{equation}
\begin{aligned}
     L_{adv} = \mathbb{E}[\log \disB{}(I_B)] +  \mathbb{E}[1 - \log \disB{}(I'_B)))] \\
    + \mathbb{E}[\log \disA{}(I_A)] +  \mathbb{E}[1 - \log \disA{}(I'_A)))]
\end{aligned}
\end{equation}
we have two discriminators \disA{}, \disB{} for distinguishing between real and generated image for A and B domain separately.

$L_{idt}$ makes sure the decoder is able to reconstruct it based on extracted content matrix and texture latent vector:
\begin{equation}
    L_{idt} = \mathbb{E}[||I_{BB} - I_B||_1] + \mathbb{E}[||I_{AA} - I_A||_1],
\end{equation}
where $$I_{AA}=\decoderA{}(T_A, \contentencodername{}(I_A)), I_{BB}=\decoderB{}(T_B, \contentencodername{}(I_B)).$$
$L_{rec}$ is based on Cycle-Consistency loss~\cite{CycleGAN2017} that optimizes the training for this under-constrained problem and regularize the translated image to preserve semantic structure of the input image:
\begin{equation}
    L_{rec} = \mathbb{E}[||I'_{BA} - I_A||_1] + \mathbb{E}[||I'_{AB} - I_B||_1],
\end{equation}
where 
$$ I'_{BA}=\decoderA{}(T_A, \contentencodername{}(I'_B)), I'_{AB}=\decoderB{}(T_B, \contentencodername{}(I'_A)).$$

To better preserve the structural information, we use perceptual loss $L_{f}$~\cite{johnson2016perceptual} based on VGG19~\cite{simonyan2014very} between the generated outputs and original Inputs. We use KL divergence loss $L_{kl}$~\cite{huang2018munit,DRIT} to minimize the distribution variance between extracted texture vectors from the target image and generated image.

Our final objective function is:
\begin{equation}
     L_{all} = L_{adv} + \lambda_{idt}L_{idt} + \lambda_{rec}L_{rec} + \lambda_{kl}L_{kl} + \lambda_{f}L_{f},
\end{equation}
where $\lambda_{adv}$, $\lambda_{idt}$, $\lambda_{rec}$, $\lambda_{kl}$ and $\lambda_{f}$ are weights assigned for each loss, respectively.

\section{Experiments}
\label{sec:experiments}

\subsection{Experimental Setup}
During training, we use Adam with 0.0005 learning rate, $\beta_{1} = 0.5$ and $\beta_{2}=0.999$. We augment the inputs and resize them to $288\times288$. Empirically, we find the model starts to converge after 600 iterations and become stable after 800 iterations for every input pair. We use widely-used dataset~\cite{CycleGAN2017} as well as images collected from the Internet (including VanillaCake$\leftrightarrow$ChocolateCake, Milk$\leftrightarrow$BubbleMilk, etc.).
We will release our Pytorch~\cite{paszke2019pytorch} implementation and associated data to facilitate future research.

\subsection{Low-level Evaluation on Augmented Data}
\begin{figure}
    \centering
	\subfigure[VanillaCake $\leftrightarrow$ ChocolateCake.] {
		\includegraphics[width=0.46\linewidth]{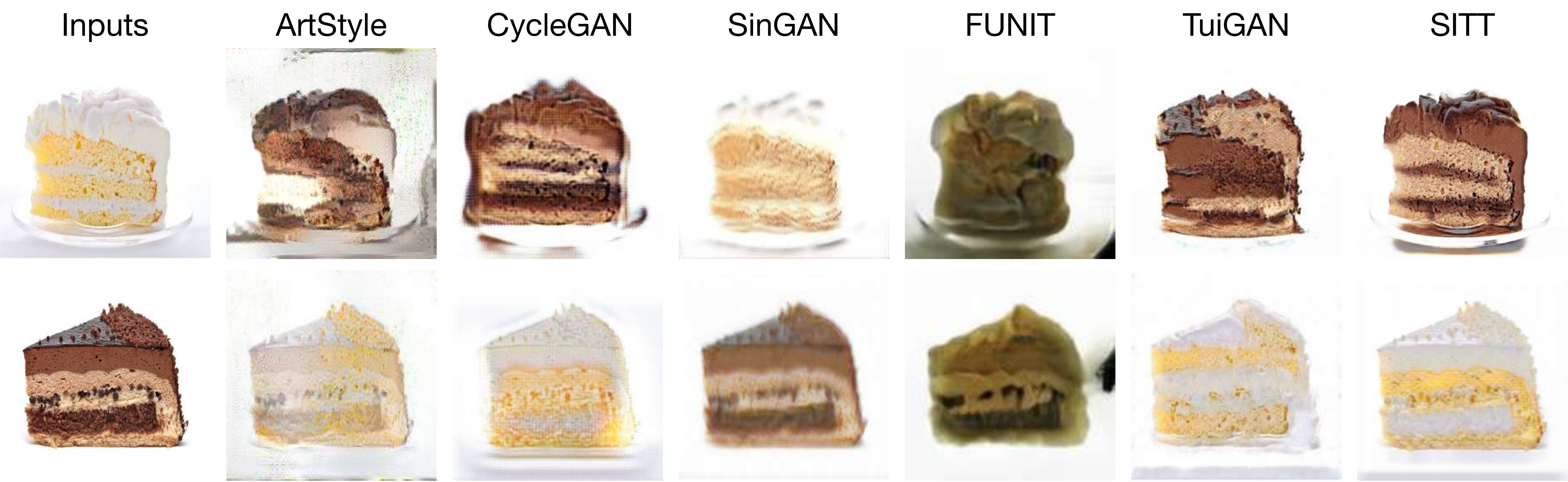}
	}
	\subfigure[Milk $\leftrightarrow$ BubbleMilk.] {
		\includegraphics[width=0.46\linewidth]{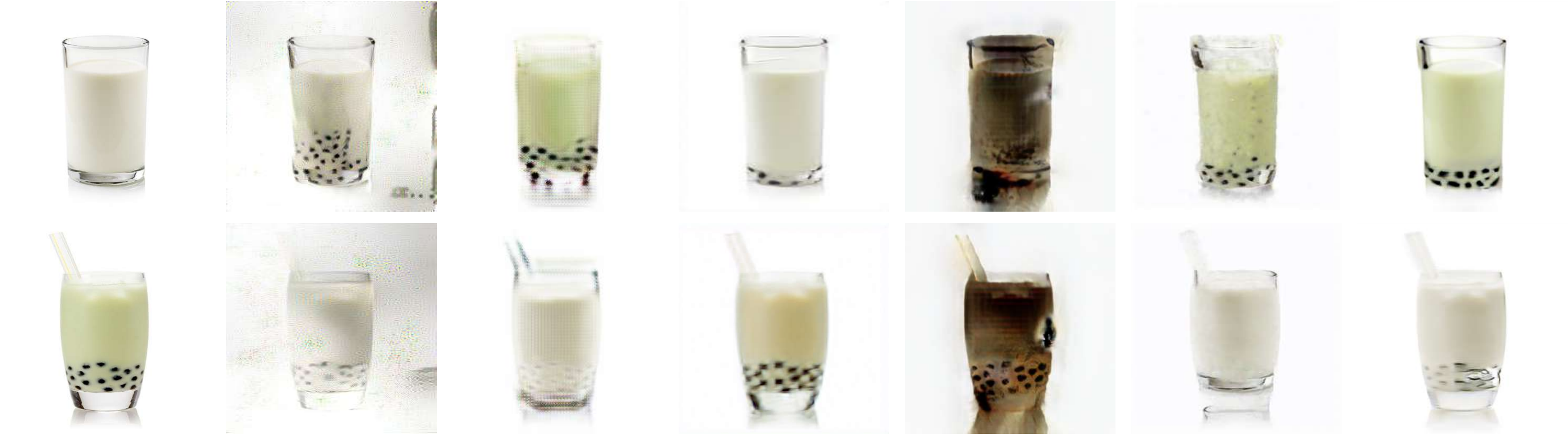}
	}
	
    \vspace{-0.1in}
    \caption{Qualitative comparison between \methodname{} and various image synthesis methods.}
    \label{fig:qualitative}
    \vspace{-0.2in}
\end{figure}

We refer to the experimental setup of TuiGAN~\cite{lin2020tuigan} that randomly selects 8 unpaired images, and generate 8 translated images for 4 unpaired image to image translation tasks including Horse$\leftrightarrow$Zebra, Apple$\leftrightarrow$ Orange,
Milk $\leftrightarrow$ BubbleMilk, and VanillaCake$\leftrightarrow$ChocolateCake. We train and test with a single source image and target texture image. We compare our results with various popular or most recent image synthesis methods with their official code and setting:  ArtStyle~\cite{gatys2015neural}, CycleGAN~\cite{CycleGAN2017}, SinGAN~\cite{shaham2019singan}~\footnote{We first train a SinGAN model and then apply it to Paint to Image.}, FUNIT~\cite{liu2019few} and TuiGAN~\cite{lin2020tuigan}. For evaluation, we use three of most popular metrics: Fréchet Inception Distance (FID)~\cite{heusel2017gans}, Learned Perceptual Image Patch Similarity (LPIPS)~\cite{zhang2018perceptual} and VGG Loss (perceptual loss)~\cite{johnson2016perceptual}. FID aims to capture the similarity of generated images to real ones, LPIPS is used to estimate how likely the outputs are to belong to the target domain, VGG Loss (with VGG19) is used to estimate how much the outputs preserve the structural information in the inputs. We randomly select 32 images and run the experiments for 3 times with a single source and target image, we report the average score in Table~\ref{tab:quantitative_1}. We could notice that \methodname{} could achieve comparatively better scores. In Fig~\ref{fig:qualitative}, we show some of the corresponding qualitative comparison results of VanillaCake $\leftrightarrow$ ChocolateCake and Milk $\leftrightarrow$ BubbleMilk. We could observe that \methodname{} enables clear and reasonable output. 
However, ArtStyle fails to learn the bubble or cake patterns, CycleGAN works better while losing detailed structural or textural information, FUNIT generates unnatural images, SinGAN doesn't change the textures appropriately and TuiGAN fails to generate reasonable samples based on multiple scales, leading to unreasonable or distorted outputs. Instead, \methodname{} doesn't depend on multi-scale training, which is more robust for random inputs of different resolutions. We also display other qualitative comparison in Appendix Section \textcolor{red}{D.1}.

\begin{table}[t]
    \centering
{
\begin{tabular}{c|c|c|c}
\toprule 
Method&FID $\downarrow$&LPIPS $\downarrow$ & VGG Loss $\downarrow$ \\  
    
\midrule\midrule
ArtStyle&237.8&0.682&0.972 \\
CycleGAN&209.6&0.514&0.805 \\
SinGAN&224.5&0.537&0.819 \\
FUNIT&221.8&0.567&0.698 \\
TuiGAN&223.7&0.513&0.791 \\
\textbf{\methodname{}}&\textbf{197.7}&\textbf{0.509}&\textbf{0.391} \\
\bottomrule
\end{tabular}
}
\caption{Quantitative comparison between \methodname{} and various image synthesis methods. The lower the better.}
\label{tab:compare_longtail_banana}
    \label{tab:quantitative_1}
    \vspace{-0.2in}
\end{table}

\begin{figure}
    \centering
    \subfigure[Similar structural modes.] {
		\includegraphics[width=.45\linewidth]{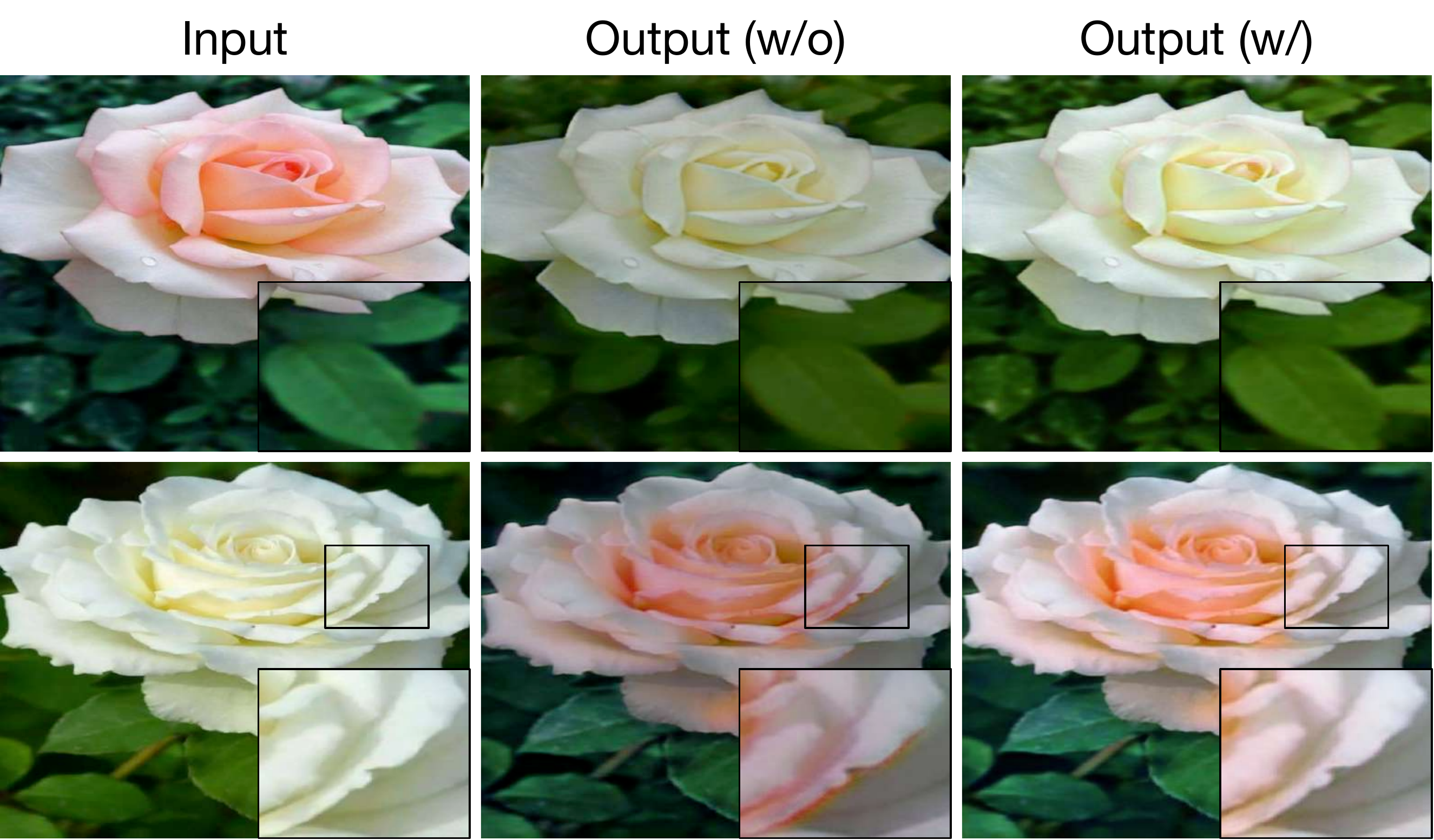}
		
	}
    \subfigure[Different structural modes.] {
		\includegraphics[width=.45\linewidth]{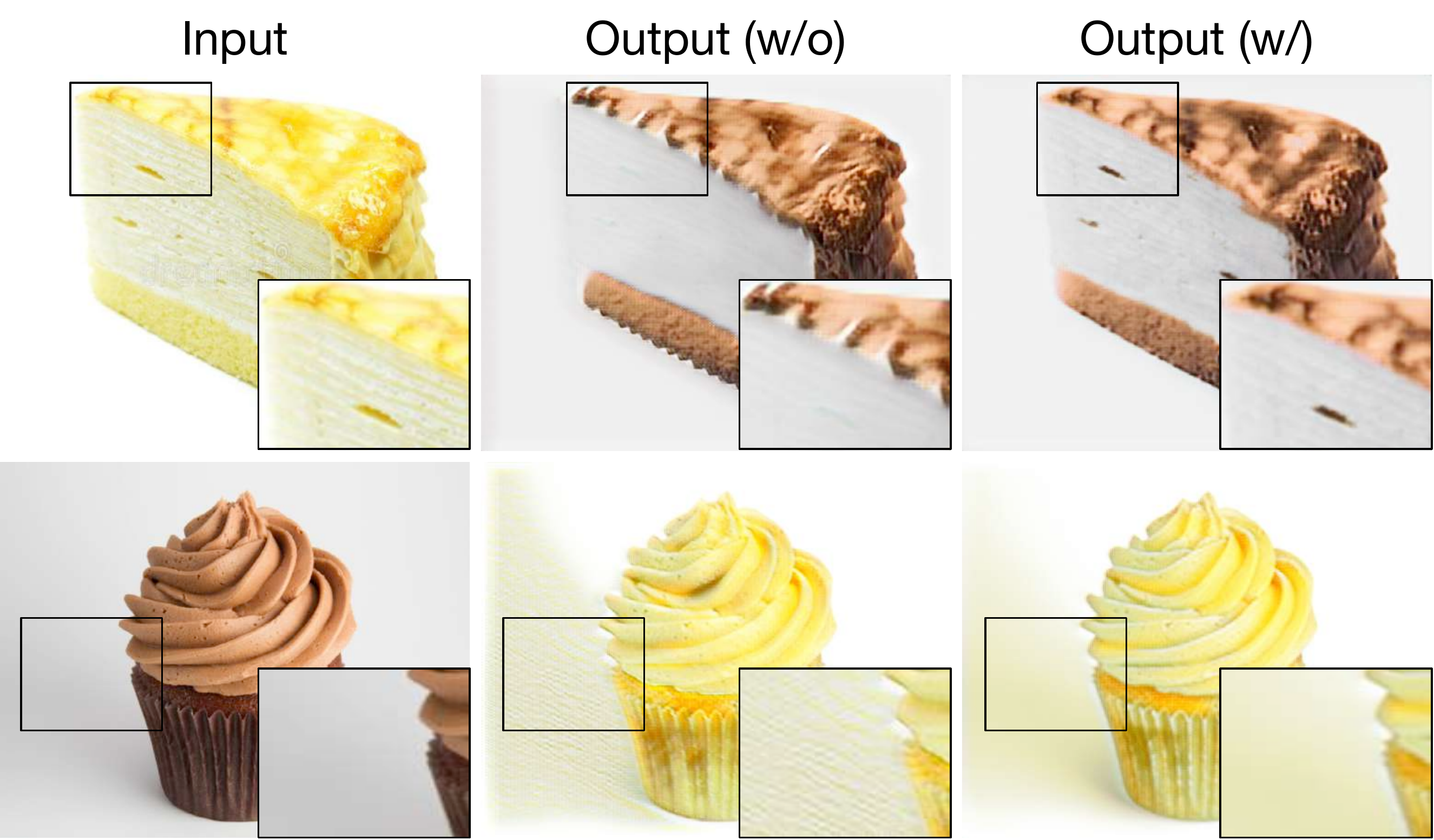}
	}
    \vspace{-0.2in}
    \caption{Compare w/ or w/o Positional Norm re-injection.}
    \label{fig:ablation_1}
    \vspace{-0.15in}
\end{figure}

\textbf{Ablation Study.}
Given that \methodname{} aims to learn a texture mapping between source image and target image but preserve general content, we utilize Positional Norm to extract structural information and re-inject them into intermediate layers of decoder. To evaluate the impact of this operation, we compare our results with or without Positional Norm re-injection. In Fig.~\ref{fig:ablation_1}, we show comparisons of similar structural modes and different structural modes that two inputs share similar or different structures. It could be noticed that Positional Norm re-injection effectively regularizes the model to preserve its original content and avoid obvious unclear regions, color distortion and weird spots for an reasonable output.

\begin{table}[t]
    \centering
{
\begin{tabular}{c|c|c|c|c}
\toprule 
Method & Testing Time $\downarrow$& Training Time$\downarrow$& Training Iters$\downarrow$ &Output Type \\
\midrule\midrule
ArtStyle&0.031&\textbf{0.036}&1000&single \\
CycleGAN&0.053&0.260&\textbf{800}&pair \\
SinGAN&0.047&0.099&20000&single \\
FUNIT&0.080&-&pre-train&pair \\
TuiGAN&0.051&0.372&20000&pair \\
\textbf{\methodname{}}&\textbf{0.009}&0.250&\textbf{800}&pair \\
\bottomrule
\end{tabular}
}
\caption{Comparisons of testing time and training time (seconds per iteration) and corresponding training details.}
\label{tab:compare_time}
    \label{tab:running_time}
    \vspace{-0.4in}
\end{table}

\textbf{Running Time.}
\methodname{} is a lightweight yet efficient network. Here we show the comparison of average training and testing time per iteration in an epoch using the same single Geforce GTX 1080 Ti GPU without acceleration in Table~\ref{tab:compare_time}. For testing, we report the time of forward operation. For training, we report the time of forward, backward and optimizer, scheduler update operations. We report the number of training iterations (iters) needed for each method.\footnote{Note: We utilize the official code or the widely used github code. For CycleGAN, we train it for the same number of epochs of \methodname{} to make fair comparisons. For FUNIT, we test the one-shot translation using the offical pre-trained model that has been trained for 100,000 iterations, so the `training time' is `-', `training iters' is `pre-train'. We report the training time of SinGAN and TuiGAN at Scale=0, which costs less time than other scales.} For output type, we report whether the model generates a pair of outputs (`pair') or a single output (`single'). The implementation of various methods are based on Pytorch~\cite{paszke2019pytorch}. We set image size as $288\times288$ in all code for fair comparison. We could notice that \methodname{} achieves fastest testing with an obvious edge. Though ArtStyle could be trained very fast, it cannot achieve texture translation very well. While \methodname{} shows its superior advantage considering training iterations, training time (seconds/iter) as well as model performance.

\subsection{Augmented Data for Image Classification}

\begin{wraptable}{R}{0.5\linewidth}
    \centering
{
\begin{tabular}{c|c|c}
\toprule 
Training data & ResNet-18 & VGG16 \\
\midrule\midrule
Baseline&55.2&55.2 \\
+ Repeat&65.7&64.6 \\
+ \methodname{}&\textbf{70.7}&\textbf{71.3}\\
\bottomrule
\end{tabular}
}
\caption{Healthy / sick leaves classification results (Top-1 accuracy \%) with different training dataset.}
\label{tab:compare_longtail_leave}
    \vspace{-0.2in}
\end{wraptable}

\subsubsection{Long-Tailed Image Classification}
\label{sec:explorations_longtailed}

\begin{figure}
    \centering
    \subfigure[\methodname{} visualization. ] {
		\includegraphics[width=.45\linewidth]{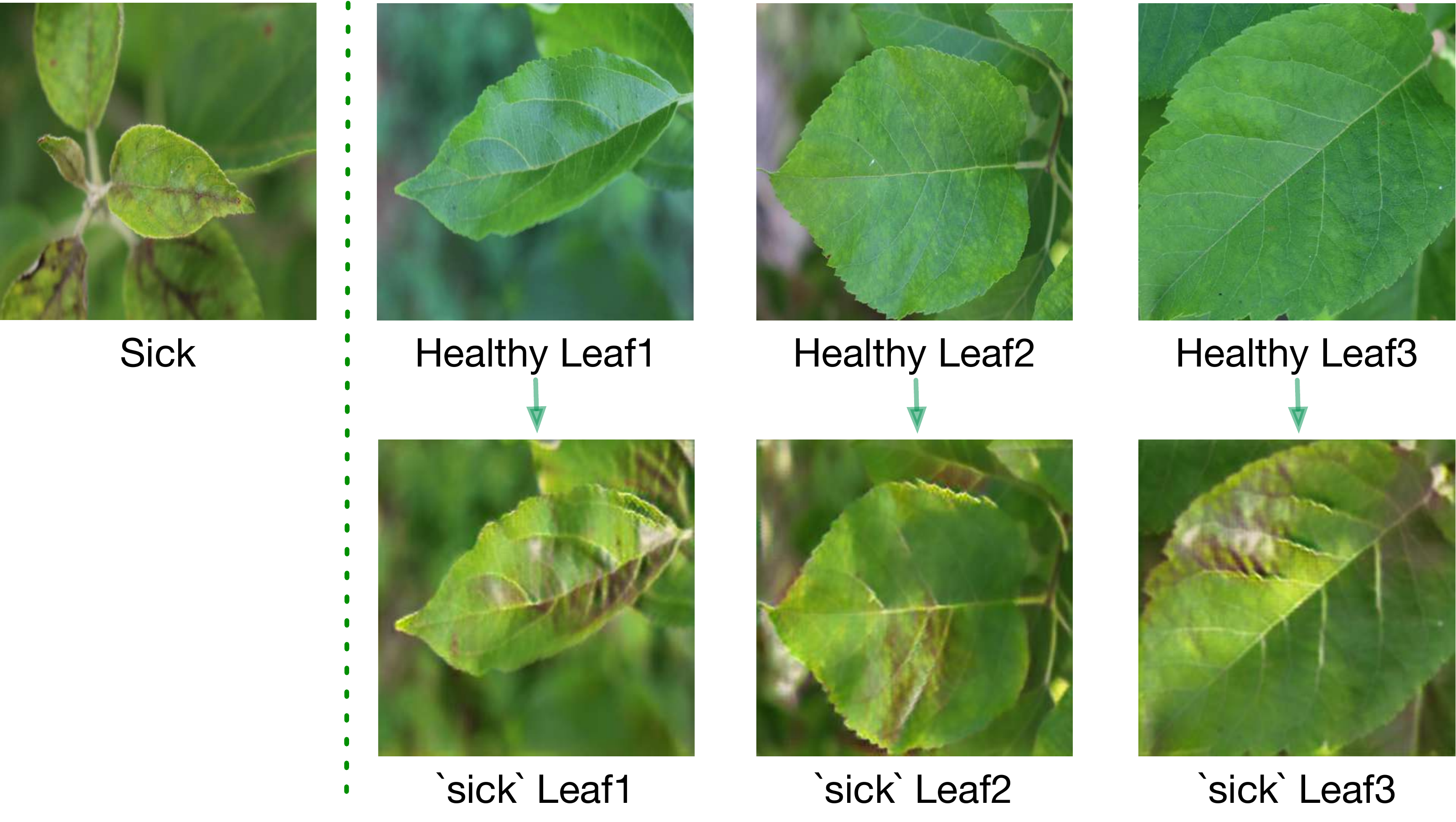}
			\label{fig:augE_leave_vis}
		
	}
    \subfigure[Augmented data visualization (multiple classes). ] {
		\includegraphics[width=.45\linewidth]{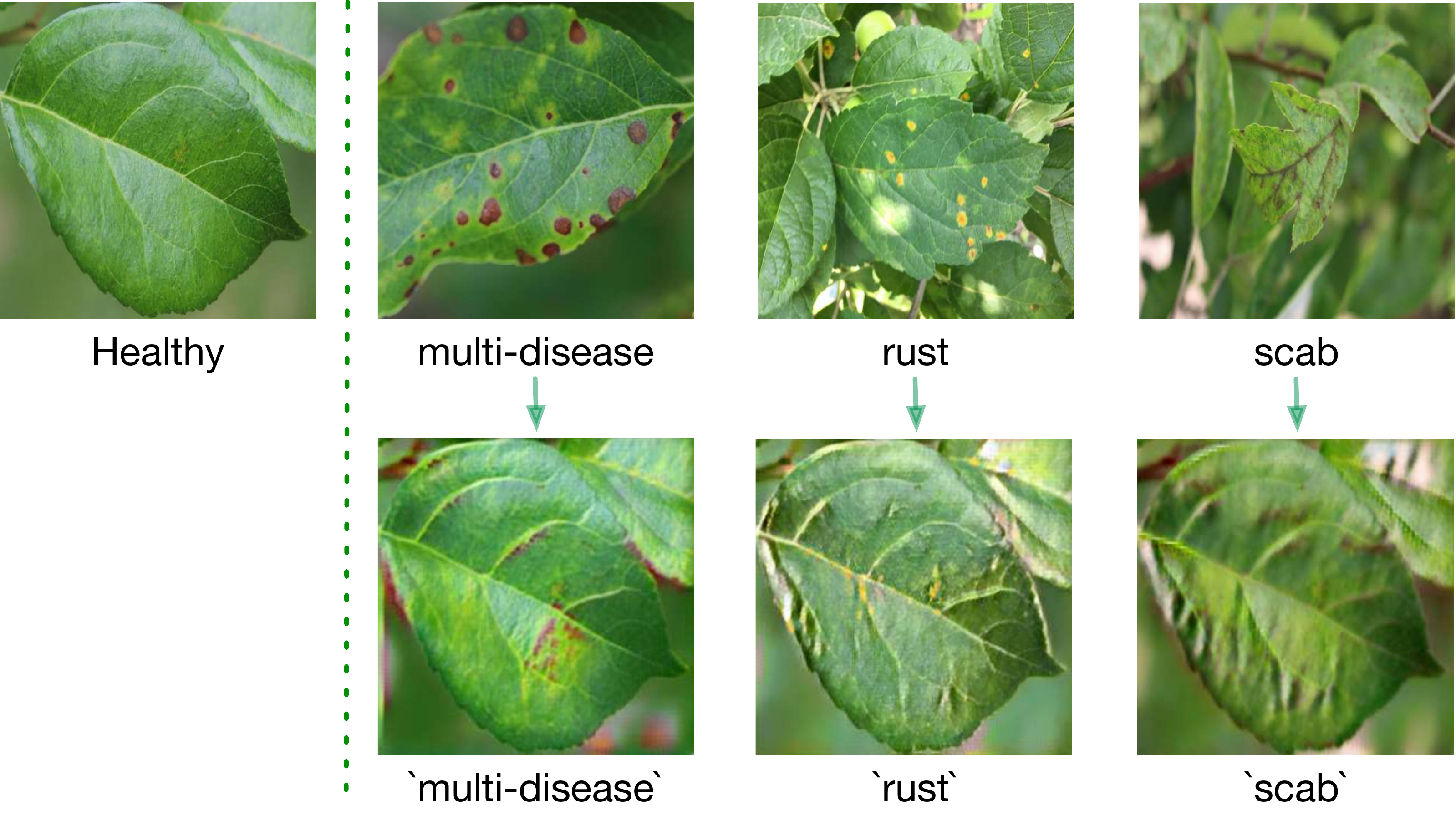}
		\label{fig:augE_leave_vis_multi}
	}
    \vspace{-0.2in}
    \caption{Compare w/ or w/o Positional Norm re-injection.}
    \vspace{-0.2in}
\end{figure}

\begin{wrapfigure}{R}{0.5\linewidth}
    \centering
    \includegraphics[width=\linewidth]{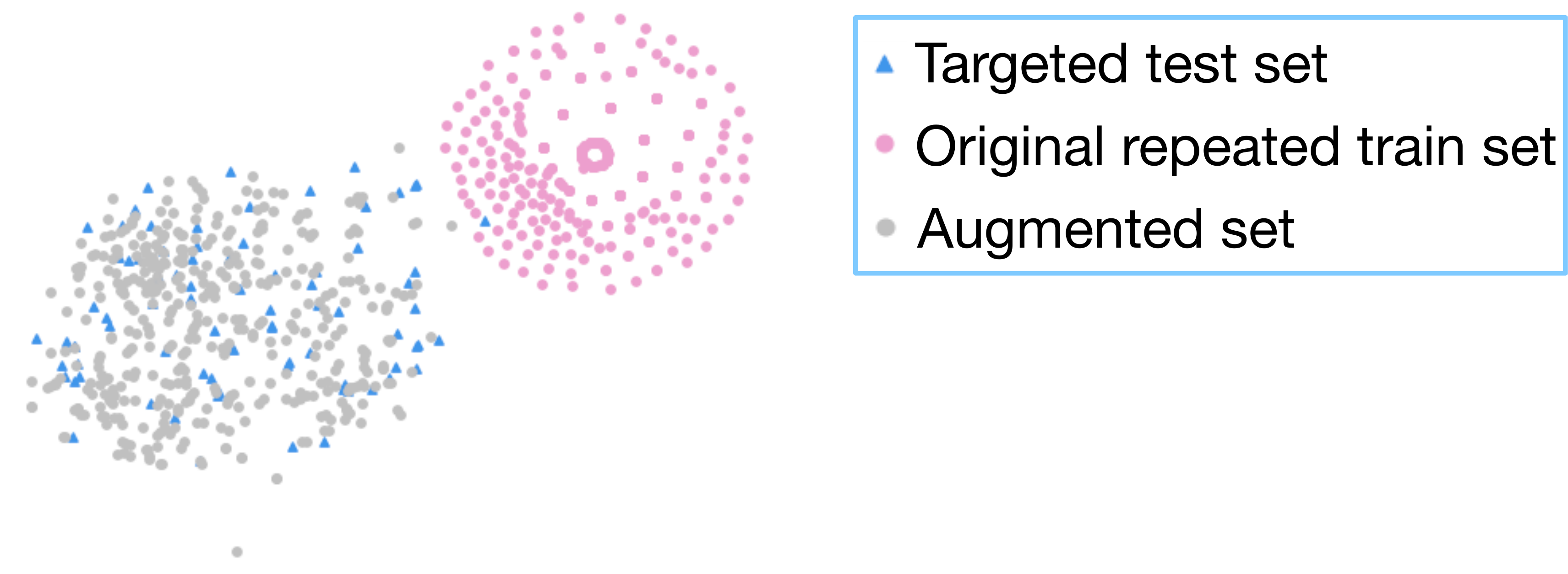}
    \caption{t-SNE visualization with leaf images.}
    \label{fig:augE_leave_tsne}
    \vspace{-0.2in}
\end{wrapfigure}

In the wild, it is very hard and impractical to collect balanced datasets for training recognition models. For example, collecting healthy leaves is easy, while collecting sick leaves could be comparatively very difficult and expensive. In this section, we use \methodname{} to translate the texture from the few sick leaves to the healthy leaves to obtain more `sick' data. We use Plant Pathology 2020 dataset~\cite{thapa2020plant} that provides data covering a number of category of foliar diseases in apple trees. We select `healthy' and `multi-disease' classes as `healthy' and `sick' and randomly split the dataset for training and testing. The training set consists of 416 healthy images and 1 sick image, the test set consists of 100 healthy images and 81 sick images. On the whole, we have three types of training data. We refer to the single sick image as \emph{Baseline}. To solve the imbalanced data problem, We repeat the single sick image to match the `healthy' data count, and refer to this operation as \emph{Repeat}. Also we apply the texture from the single sick leaf to all healthy leaves via \methodname{} (Please see Fig~\ref{fig:augE_leave_vis} for examples). To validate the augmented dataset, we use t-SNE~\cite{maaten2008visualizing} to visualize the distribution of Repeat, targeted test set, and \methodname{}. In Fig.~\ref{fig:augE_leave_tsne}, we see that the augmented data shares a similar t-SNE distribution with the targeted test set. 

We train and test with these datasets on ResNet-18~\cite{he2016deep} and VGG16~\cite{simonyan2014very} using cosine learning rate starting from 0.01 for 90 epochs
with standard augmentation RandomResizedCrop and RandomHorizontalFlip. We run experiments for three times and report the average score for fair comparison. In Table~\ref{tab:compare_longtail_leave}, we could observe consistent superior results of \methodname{} over the baselines. 

\begin{figure}[ht]
    \centering
    \includegraphics[width=0.8\linewidth]{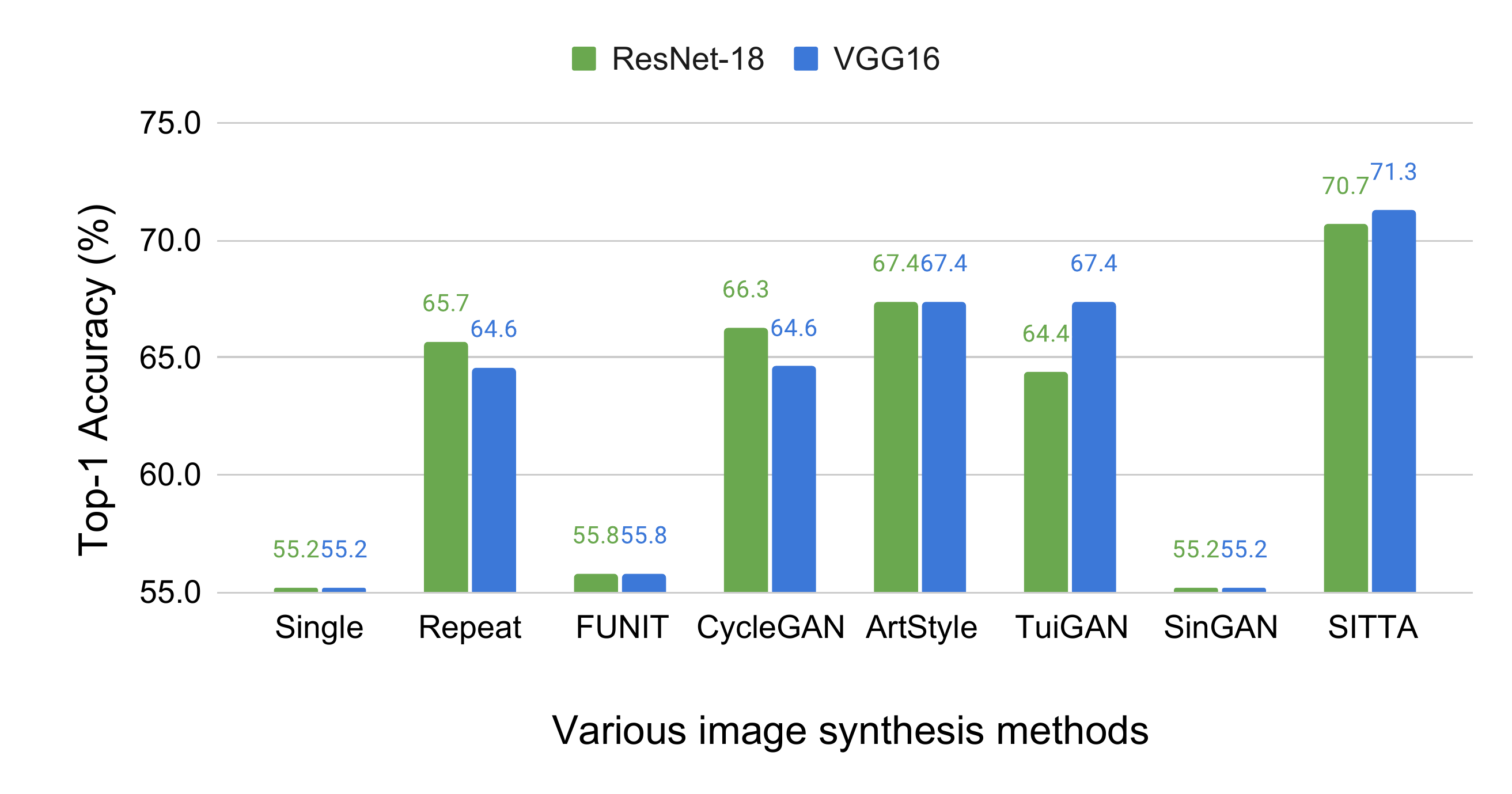}
    \vspace{-0.3in}
    \caption{Healthy / Sick leaves classification results (Top-1 accuracy \%) with different image synthesis methods.}
    \label{fig:compare_synthesis_methods_longtail_leave}
    \vspace{-0.2in}
\end{figure}

\textbf{Comparison with various image synthesis methods for data augmentation.}
We emphasize that it is not easy to synthesize useful data for image recognition. To further clarify the concerns, based on Table~\ref{tab:compare_longtail_leave} setting, we conduct an ablation study based on Healthy / Sick leaves classification generated by various image synthesis methods. Please note that most methods cannot be finished within a short time and is impractical for large-scale data augmentation as shown in Table~\ref{tab:compare_time}. We display the results in Fig~\ref{fig:compare_synthesis_methods_longtail_leave}. We could observe that \methodname{} significantly outperforms other methods for data augmentation. 

\textbf{Complementarity with other augmentation methods.} 
\methodname{} is a simple yet efficient method to generate new data semantically, and we may regard it as complimentary with other existing augmentation methods. To justify this, we further explore leaf classification in the multiple classes setting. We add the other classes `rust' and `scab' from Plant Pathology 2020~\cite{thapa2020plant}, aiming to classify which category the leaves belong to. We randomly split the dataset for training and testing. The training set consists of 416 healthy leaf images, 1 multi-disease, 1 rust and 1 scab leaf image, the test set consists of 100 healthy, 81 multi-disease, 612 rust and 582 scab leaf images. We take all four categories for 4-class classification and healthy, rust, scab leaf images for 3-class classification, as well as healthy, multi-disease for 2-class classification. Same as previous setting, we treat the original training data as baseline. We augment the training data via Repeat and \methodname{} respectively. We display \methodname{} augmented examples in Fig~\ref{fig:augE_leave_vis_multi}. Since we witnessed no improvement for most data augmentation methods for only a single input, we compare our results with more competitive settings: using different data augmentation methods based on repeated data. We combine Repeat and \methodname{} with different widely-used and recent augmentation methods~\cite{chen2020improved} including Colorjitter, GaussianBlur, Grayscale, Mixup~\cite{zhang2018mixup}, CutMix~\cite{yun2019cutmix} and MoEx~\cite{li2021feature} using the official recommended hyper-parameters for ResNet. In Table~\ref{tab:compare_multi_classes_1}, we show the comparison of various augmentation strategies based on ResNet-18. Consistent with previous results, \methodname{} improves the baseline with a large margin and is even highly competitive with state-of-the-art methods Mixup, CutMix and MoEx. Also, since \methodname{} generates new data that could be easily used with any augmentation methods. It could be observed that \methodname{} is compatible with other augmentation methods and consistently help boost the model performance with obvious edge.

\vspace{-0.15in}

\begin{table}[t]
    \centering
{
\begin{tabular}{l|c|c|c}
\toprule 
Method & 2-class&3-class & 4-class \\
\midrule\midrule
Baseline&55.2&8.4&7.8 \\
\midrule
+ Repeat&61.9&23.4&28.8 \\
+ Repeat + GaussianBlur&63.0&23.7&29.6\\
+ Repeat + Grayscale&65.2&28.9&32.9\\
+ Repeat + Colorjitter&64.6&\textbf{35.5}&35.5\\
+ Repeat + Mixup&60.2&17.5&20.1\\
+ Repeat + CutMix&65.2&27.0&26.7\\
+ Repeat + MoEx&60.8&26.8&35.6\\
\midrule
+ \methodname{}&70.7&27.4&30.7 \\
+ \methodname{} + MoEx&\textbf{72.9}&32.3&\textbf{37.5}\\
\bottomrule
\end{tabular}
}
\caption{Multi-class leaves classification results (Top-1 accuracy \%, average of 3 runs) with various augmentation strategies based on ResNet-18.}
\label{tab:compare_longtail_multi_classes}

    \label{tab:compare_multi_classes_1}
    \vspace{-0.2in}
\end{table}

\subsubsection{Few-shot Image Classification}
\label{sec:explorations_fewshot}
Modern recognition systems are data-intensive and often need many examples of each class to saturate performance.
However, it is impractical and hampered when the data set is small. Few-shot learning~\cite{wang2018low,wang2020generalizing} is proposed to solve such kind of problem and improve the recognition performance by training with few samples. In this section, \methodname{} aims to generate additional training images for improving few-shot image classification. 

\textbf{Oxford 102 flowers.} Here we set up all-way few-shot (few images of all classes) classification based on Oxford 102 flowers dataset~\cite{nilsback2008automated} that consists of 102 flower categories, each class consists of between 40 and 258 images. We strictly follow the official data split rule. For training with \methodname{}, we randomly select 5 images as 5 shots for each class (Baseline), for testing, we evaluate the model on all test images. We augment the data within each category, each image is augmented to 4 additional images based on the new content of other 4 images (Please see Appendix Section \textcolor{red}{D.3} for a bunch of visualization examples), therefore we have 25 images for each class. To relieve the concern of imbalanced class, we repeat 5 images of each category for 4 times to ensure the same numbers of train set. Since all-way 5-shot 102 classes classification is very difficult to train from scratch~\cite{Sun_2019_CVPR}, we fine-tune the pre-trained model and test on the official testset based on ResNet-18, VGG16 and ResNet-50 (all of which have been pretrained on the 1000-class ImageNet~\cite{deng2009imagenet}). For fine-tuning, we fine-tune the Pytorch default pre-trained model for 90 epochs and set cosine learning rate starting from 0.01. We run the experiments for 3 times and report the average score in Table~\ref{tab:compare_fewshot_102flowers}. It could be obviously noticed that \methodname{} is able to improve the classification on 102 classes without any additional fine-tuning on real images. The results are very inspiring and show the consistent edge regarding different model architectures. 

\begin{table}[t]
    \centering
{
\begin{tabular}{c|c|c|c}
\toprule 
Method & ResNet-18 & VGG16& ResNet-50 \\
\midrule\midrule
Baseline&73.8&75.0&79.6\\
+ Repeat&76.2&	77.7&	81.8 \\
+ \methodname{}&\textbf{77.7}&\textbf{79.2}&\textbf{82.7}\\
\bottomrule
\end{tabular}
}
\caption{Comparison of all-way 5-shot classification (Top-1 accuracy \%) results on Oxford 102 flowers.}

    \label{tab:compare_fewshot_102flowers}
    \vspace{-0.4in}
\end{table}

\begin{table}[t]
    \centering
{
\begin{tabular}{c|c|c|c}
\toprule 
Method & ResNet-18 & VGG16& ResNet-50 \\
\midrule\midrule
Baseline&30.9&38.9&38.3\\
+ Repeat&31.1&	40.4&	38.6\\
+ \methodname{}&\textbf{32.5}&\textbf{40.5}&\textbf{38.9}\\
\bottomrule
\end{tabular}
}
\caption{Comparison of all-way 5-shot classification (Top-1 accuracy \%) results on CUB-200-2011.}

    \label{tab:compare_fewshot_CUB200}
    \vspace{-0.4in}
\end{table}

\textbf{Caltech-UCSD Birds 200.} To further verify the validity of \methodname{}, we strictly follow the same procedure of Oxford flowers setting and apply \methodname{} to Caltech-UCSD Birds-200-2011 (CUB-200-2011) dataset~\cite{wah2011caltech} for data augmentation (Please see Appendix Section \textcolor{red}{D.4} for a gallery of visualization examples). CUB-200-2011 consists of 11,788 photos of 200 bird species. We run the experiments for 3 times and report the average classification score on the official testset in Table~\ref{tab:compare_fewshot_CUB200}. We observe the consistent and competitive improvement of \methodname{}, which sheds light on the feasibility of \methodname{} for few-shot learning tasks.

\textbf{iNaturalist Birds.}
iNaturalist (iNat)~\cite{van2018inaturalist} is a large-scale species classification and detection dataset that features visually similar species, captured in a wide variety of situations from all over the world. We select two genera House Sparrow and Parulidae under the Aves (or bird) supercategory from iNat 2018 (training images) that provides data and labels based on biology taxonomy. We randomly split the dataset into train and test. The training set consists of 518 Parulidae images and 1 House Sparrow image, both test set consists of 105 images separately. Since the two categories share similar structures, we are able to use \methodname{} to generate synthetic images with translated textures (Please see Appendix Section \textcolor{red}{D.2} for illustration). Following previous long-tailed classification settings, we compare the classification results and display them in Table~\ref{tab:compare_longtail_bird}. It could be observed that augmented birds are able to help improve the classification performance, which is consistent with the results of leaf classification. 

\begin{wraptable}{R}{0.5\linewidth}
    \centering
{
\begin{tabular}{c|c|c}
\toprule 
Training data & ResNet-18 & VGG16 \\
\midrule\midrule
Baseline&50.0&50.0 \\
+Repeat&57.1&63.8 \\
+\methodname{}&\textbf{61.4}&\textbf{68.6}\\
\bottomrule
\end{tabular}
}
\caption{Comparison of iNat bird classification results(Top-1 accuracy \%). }
\label{tab:compare_longtail_bird}
    \vspace{-0.2in}
\end{wraptable}

\vspace{-0.1in}
\section{Discussion}
\label{sec:discussion}
\begin{wrapfigure}{R}{0.5\linewidth}
    \centering
    \includegraphics[width=\linewidth]{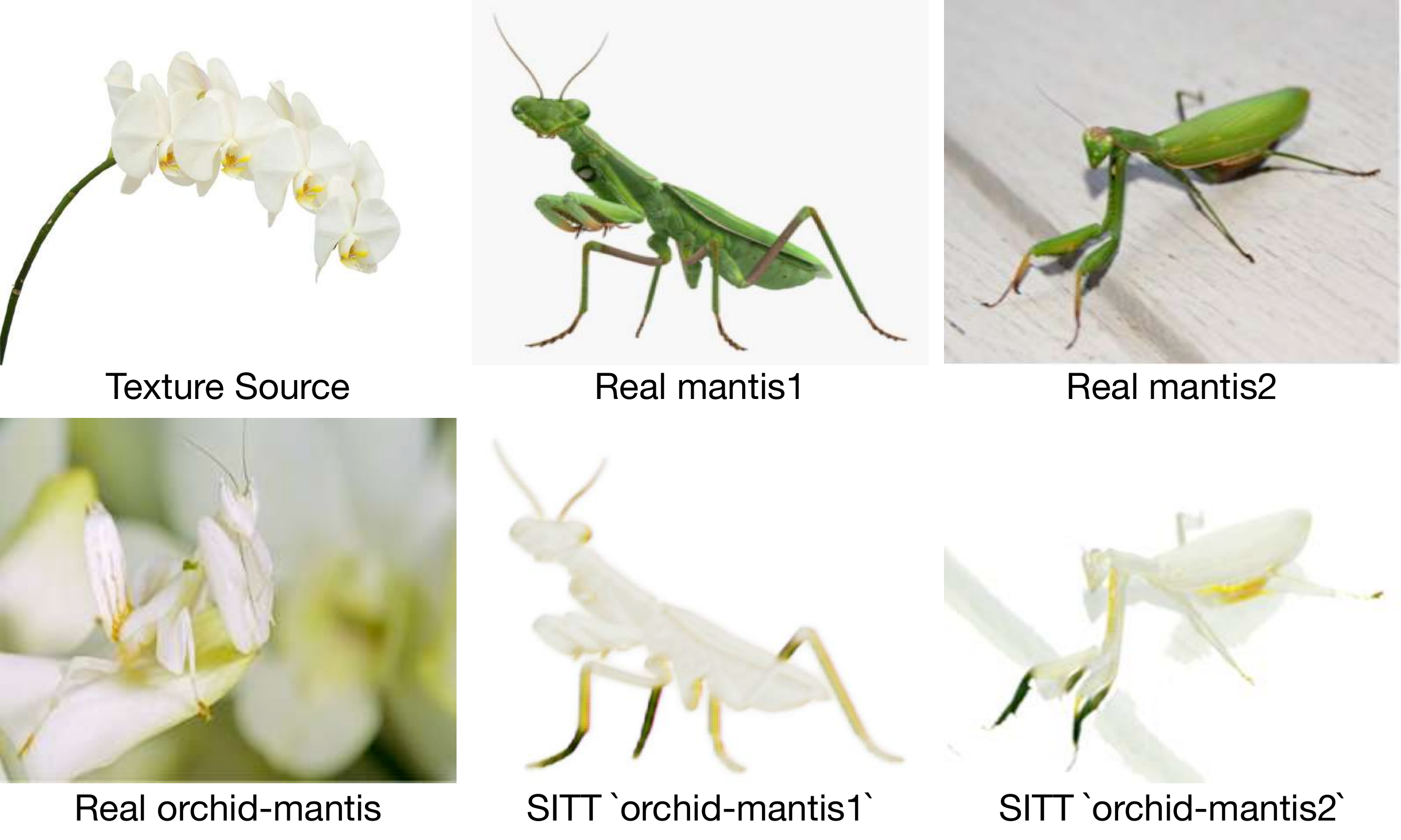}
    \vspace{-0.1in}
    \caption{Augmented Orchid-Mantis for camouflage. }
    \label{fig:augE_orchid_mantis}
    \vspace{-0.2in}
\end{wrapfigure}

\textbf{Camouflage.} 
Mimicry or Camouflage in natural world provides some real examples for texture swapping that creatures make the textures of their body similar to the environment's to avoid danger or hunt food~\cite{thery2002predator}. For instance, we show the case of mantis and orchid. To prey for the insects, mantis will adaptively change their texture similar to the orchids. In Figure~\ref{fig:augE_orchid_mantis}, we give an illustration and find \methodname{} could translate the orchids' texture to the mantis and obtain reasonable and natural outputs that looks very close to the real example. 

\textbf{Task-specific augmentation.} 
In this paper, we introduce task- and dataset-specific augmentation. We aim to swap the texture between a target object (for shape) and an exemplar object (for texture). Therefore, we don't want the shape with the new texture confuses with other classes in the dataset. 
In traditional image recognition pipelines, image augmentation includes rotation, crop, jittering, flip, etc. Recently, ~\cite{zhang2018mixup,li2021feature} revised input data or features to improve image classification performance. On the one hand, these methods change the way of representation while keeping the texture the same. On the other hand, image synthesis brings a new direction to change the texture while keeping the structure unchanged. Such kind of augmentation engineering opens a door for understanding objects from textures to structure, as well as augmenting data by `destroy' the original textures while replacing them with sensible or other resources. 

\textbf{Label assignments for augmented data.} Image synthesis brings a new world for data augmentation. Previous methods mainly focus on leading networks to learn more shape information by applying different styles to the original images~\cite{geirhos2018imagenet,zheng2019stada,somavarapu2020frustratingly}. In these cases, the augmented image will be assigned with the original label. However, label assignment should be serious considered given different tasks and datasets. For leaves classification, textures make sick and healthy leaves different. Therefore, we generate a bunch of sick leaves by translating the textures of sick leaves to healthy leaves and assign the new image as `sick' instead of `healthy.' While for few-shot learning experiments, we augment the data within each category and assign the same label to them. For future potential directions, more advanced label assignment strategies such as label perturbation~\cite{zhang2018mixup,yun2019cutmix,li2021feature} could be considered for synthetic dataset.  

\textbf{Limitations.} Although \methodname{} brings an obvious improved margin for recognition models, sometimes there are very few provided texture source images. Therefore the extracted texture information via \methodname{} might not be able to cover all texture patterns in testset, leading to limited accuracy improvement. Besides, in spite of faster running time, there still is a processing time gap between image synthesis and traditional data augmentation such as flip and crop. For the future work, we would explore on both optimally increasing the diversity based on the single or very few texture source and speeding up the image translation model. We believe these signals shed light on the new research direction of semantic image synthesis for data augmentation.

\vspace{-0.1in}
\section{Conclusion}
\label{sec:conclusion}
\vspace{-0.1in}
In this paper, we explore the problem of image synthesis for recognition tasks. We propose a lightweight, fast and efficient Single Image Texture Translation for data Augmentation (\methodname{}). Images generated by \methodname{} not only look appealing but also help visual recognition tasks including long-tailed and few-shot image classification.
We hope our work could open the door of using image synthesis for data augmentation in various computer vision tasks and make image synthesis one step closer to solving real problems in the wild.

\vspace{-0.2in}
\section*{Acknowledgement}
\vspace{-0.1in}
\label{sec:acknowledgement}
This work was supported in part by the Pioneer Centre for AI, DNRF grant number P1.
\clearpage
%
%
{\small
\bibliographystyle{ieee_fullname}
\bibliography{reference}
}
\newpage
\appendix
\label{sec:appendix}
\section{\methodname{} Workflow}
\label{appen_aug_workflow}
Assume we have plenty of content source images $SetA$ but a few target texture source images $SetB$, the only difference between $SetA$ and $SetB$ is that they contain different textures. Our goal is to synthesize extra `B` data by replacing textures of $SetA$ with the new textures from $SetB$, we call the augmented dataset as $AugSetB$. Please see Algorithm~\ref{alg:aug_workflow} for details.
\begin{algorithm}
\SetAlgoLined
\KwResult{Augmented dataset $AugSetB$}
\textbf{Input:} $SetA$, $SetB$\;
 \uIf{Single to Single}{
 \While{Image $I_A$ in $SetA$}{
 \While{Image $I_B$ in $SetB$}{
 initialize $\methodname{}$\;
 \While{not converged}{
 train $\methodname{}$\;
 }
  $I'_B \xleftarrow{} \methodname{}(I_B, I_A)$\;
  $AugSetB \xleftarrow{} AugSetB \cup I'_B $\;
   }}
   }
  \uElseIf{Single to Multi}{
   \While{Image $I_B$ in $SetB$}{
   initialize $\methodname{}$\;
 \While{Image $I_A$ in $SetA$}{
 \While{not converged}{
  train $\methodname{}$\;
  }
   }
   \While{Image $I_A$ in $SetA$}{
  $I'_B \xleftarrow{} \methodname{}(I_A, I_B)$\;
  $AugSetB \xleftarrow{} AugSetB \cup I'_B $\;
   }
   }
   
 }
 \caption{\small{\methodname{} Workflow}}
 \label{alg:aug_workflow}
\end{algorithm}

\section{\methodname{} Model Design}
\label{appen_sitt_workflow}
\methodname{} consists of three parts: shared Texture Encoder (\textureencodername{}), shared Content Encoder (\contentencodername{}) and corresponding Decoder \decoderA{}, \decoderB{}  for each category. We first feed the Input B $I_B$ into \textureencodername{} to get texture latent vector $T_B$ and feed texture source Input A $I_A$ into the \contentencodername{} to get content matrix $C_A$. Then we feed $T_B$ and $C_A$ into \decoderB{} to get $I'_B$. Translating B's texture to A follows the same procedure. To better preserve the structural information and optimize training, we apply Positional Norm in \contentencodername{} to extract intermediate normalization constants mean $\mu$ as $\beta$ and standard deviation $\sigma$ as $\gamma$ and re-inject them into the later layers of Decoder to transfer structural information. Please see Algorithm~\ref{alg:sitt_workflow} for details.

\begin{algorithm}
\SetAlgoLined
\KwResult{Translated output $I'_A$, $I'_B$}
\textbf{Input:} $I_A$, $I_B$\;
 initialize \contentencodername{}, \textureencodername{}, \decoderB{}\;
 \While{not converged}{
  Step1: $T_B \xleftarrow{} \textureencodername{}(I_B)$\; $T_A \xleftarrow{} \textureencodername{}(I_A)$\;
  Step2: $C_A, \beta_{A}, \gamma_{A} \xleftarrow{} \contentencodername{}(I_B)$\;
  $C_B, \beta_{B}, \gamma_{B} \xleftarrow{} \contentencodername{}(I_B)$\;
  Step3: $I'_B \xleftarrow{}  \decoderB{}(T_B, C_A, \beta_{A}, \gamma_{A})$\; $I'_A \xleftarrow{}  \decoderA{}(T_A, C_B, \beta_{B}, \gamma_{B})$\;
  Step4: Backpropagation\; 
 }
 \caption{\methodname{} Model Design}
 \label{alg:sitt_workflow}
\end{algorithm}

\section{\methodname{} for Few-shot Image Classification}
\label{appen_data_aug_study}
Besides Oxford 102 Category flower dataset~\cite{nilsback2008automated}, we also conduct study on 17 Category flower dataset~\cite{Nilsback06} that is composed of 17 category with 80 images for each class. The dataset has been randomly split into 3 different training, validation and test sets. We strictly follow the data split rule. Same with previous all-way 5-shot experimental setting, we randomly select 5 images in each category for training and test the model on the official testset. Based on 3 different split dataset, we run the experiments for 3 times and report the average score of three testsets in Table~\ref{tab:compare_fewshot_17flowers}. We could observe that the augmented dataset via \methodname{} still brings consistent competitive improvement.

\begin{table}[t]
    \centering
{
\begin{tabular}{c|c|c|c}
\toprule 
Method & ResNet-18 & VGG16& ResNet-50 \\
\midrule\midrule
Baseline&80.8&73.8&84.8\\
+ Repeat&82.1	&82.5&84.9	 \\
+ \methodname{}&\textbf{83.6}&\textbf{83.0}&\textbf{85.1}\\
\bottomrule
\end{tabular}
}
\caption{Comparison of all-way 5-shot classification (Top-1 accuracy \%) results on Oxford 17 flowers.}

    \label{tab:compare_fewshot_17flowers}
\end{table}

\section{More \methodname{} Examples on Natural Images}
\label{appen_data_aug_example}
\subsection{More Qualitative Comparison}
\label{appen_data_aug_example_qualitative_com}
\begin{figure}
    \centering
    \subfigure[Horse $\leftrightarrow$ Zebra.] {
		\includegraphics[width=3.5in, height=1.1in]{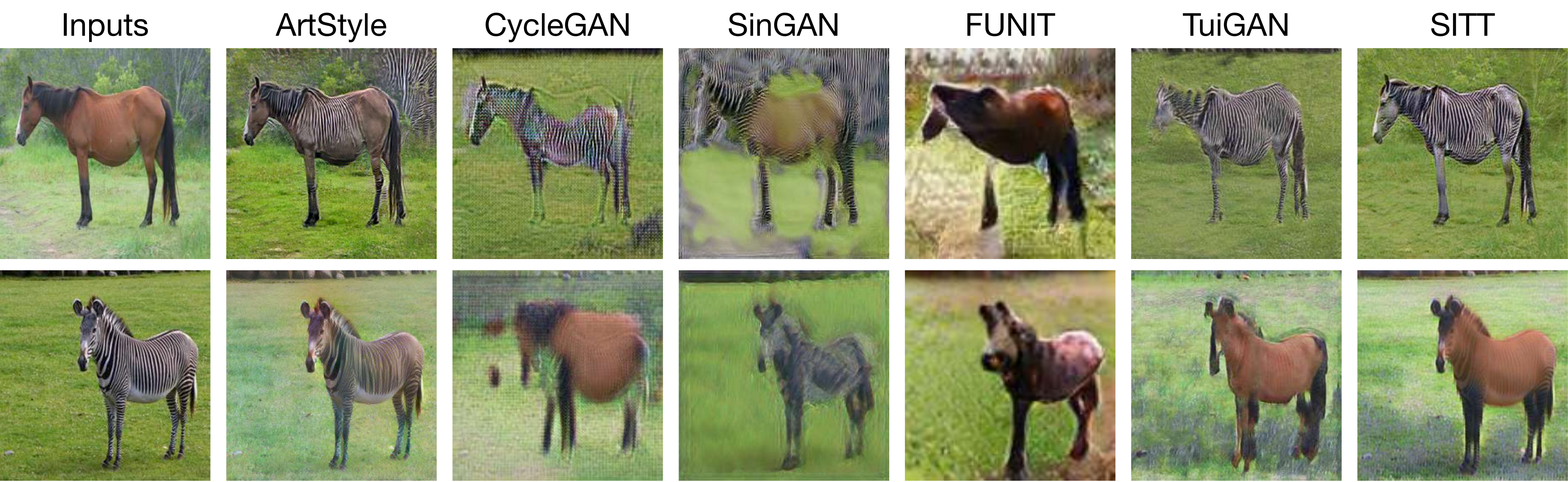}
	}
	\subfigure[Apple $\leftrightarrow$ Orange.] {
		\includegraphics[width=3.5in, height=1in]{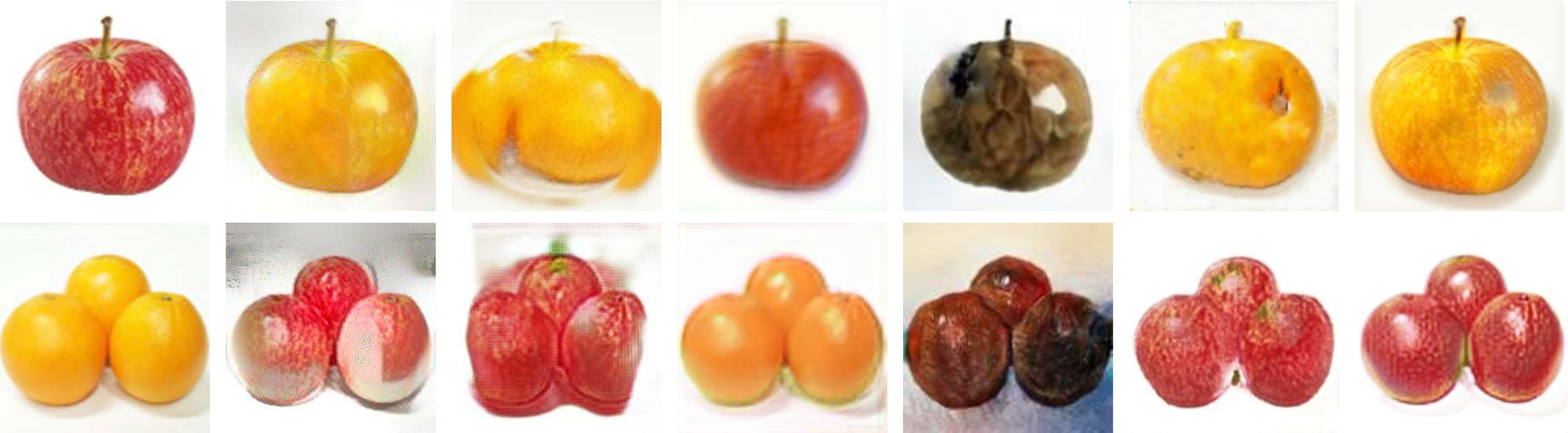}
	}

    \caption{Qualitative comparison between \methodname{} and several state-of-the-art image synthesis methods.}
    \label{fig:qualitative_appendix}
\end{figure}

In Figure~\ref{fig:qualitative_appendix}, we display more qualitative comparison with several state-of-the-art image synthesis methods.

\subsection{iNaturalist Birds}
\label{appen_data_aug_example_inat_birds}
\begin{figure}
    \centering
     \includegraphics[width=\linewidth]{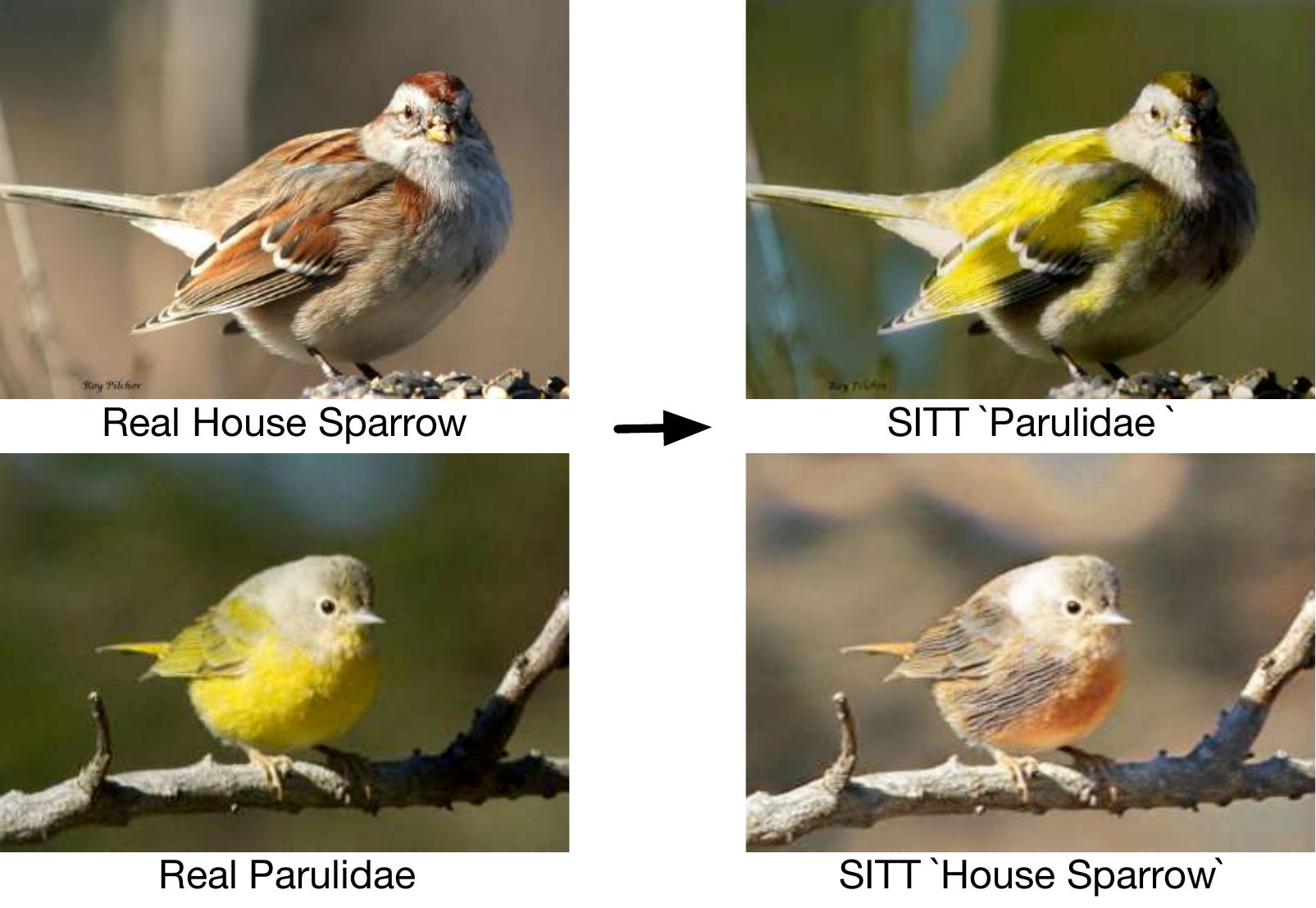}
    \caption{\methodname{} translates the texture from one image to another.}
    \label{fig:augE_bird_vis}
\end{figure}

In Figure~\ref{fig:augE_bird_vis}, we illustrate another texture swapping example between two species of birds, while preserving the natural appearances.

\subsection{Oxford 102 flowers}
\label{appen_data_aug_example_oxford_102}
\begin{figure}
    \centering
     \includegraphics[width=\linewidth]{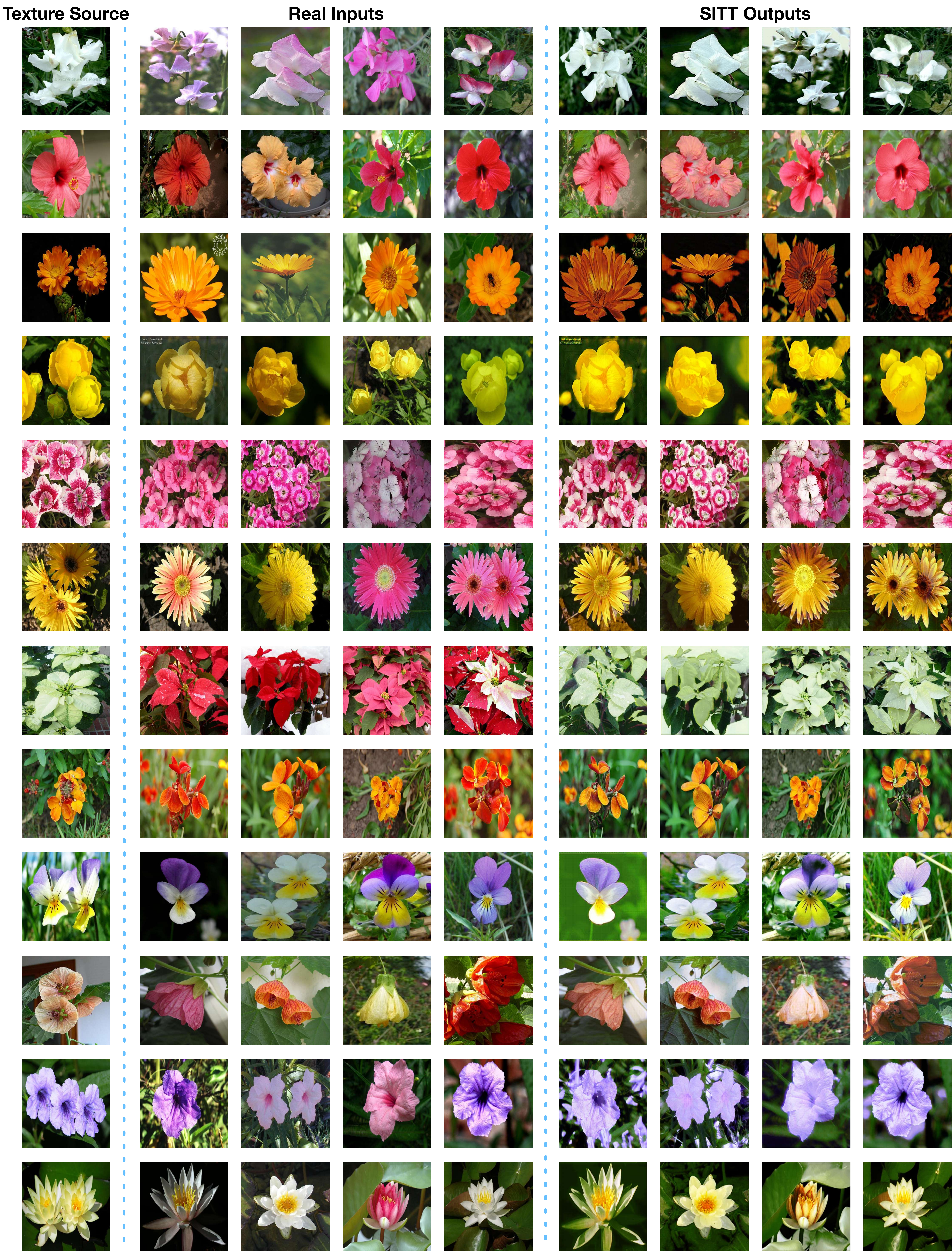}
    \caption{Augmented Oxford flowers visualization via \methodname{}. }
    \label{fig:appen_flowers_vis1}
\end{figure}

Please see Figure~\ref{fig:appen_flowers_vis1} for details.

\subsection{Caltech-UCSD Birds 200}
\label{appen_data_aug_example_cub200}
\begin{figure}
    \centering
     \includegraphics[width=\linewidth]{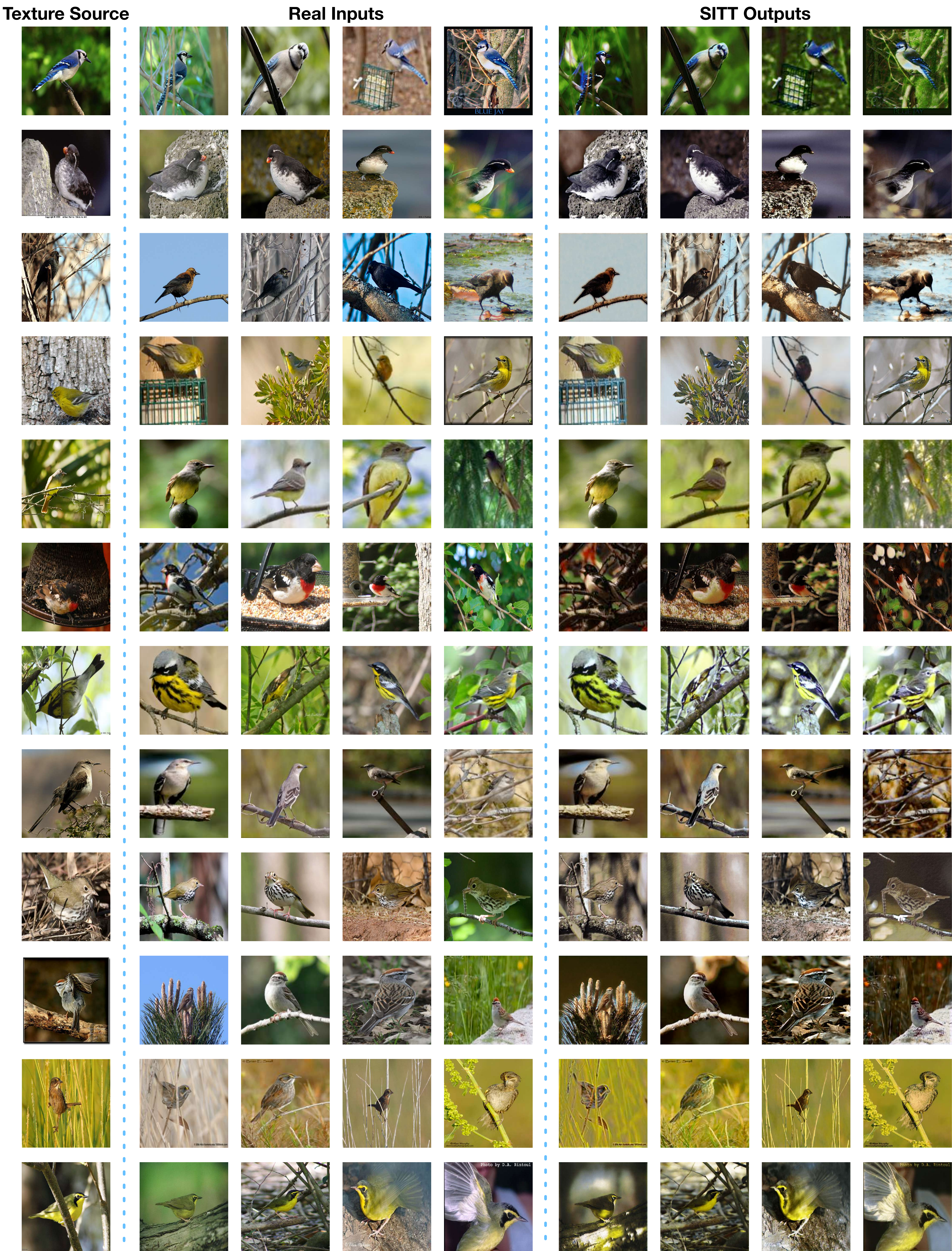}
    \caption{Augmented CUB-200-2011 visualization via \methodname{}.}
    \label{fig:appen_cub200_vis1}
\end{figure}

s
Please see Figure~\ref{fig:appen_cub200_vis1} for details.

\end{document}